\documentclass[lettersize,journal]{IEEEtran}
\usepackage{amsmath,amsfonts}
\usepackage{algorithm}
\usepackage{array}
\usepackage[caption=false,font=normalsize,labelfont=sf,textfont=sf]{subfig}
\usepackage{textcomp}
\usepackage{stfloats}
\usepackage{url}
\usepackage{verbatim}
\usepackage{graphicx}
\usepackage{cite}
\newtheorem{definition}{Definition}
\newtheorem{thm}{Theorem}

\usepackage{multicol}
\usepackage{multirow}
\usepackage{amsfonts}
\usepackage{amssymb}
\usepackage{colortbl}
\usepackage{enumitem}
\usepackage{booktabs}
\usepackage{xcolor} 
\usepackage[algo2e]{algorithm2e}

\newcommand{\baserule}{RuleTensor-TSP}

\newcommand{\model}{GPHT}
\newcommand{\new}[1]{\textcolor{black}{#1}}
\definecolor{light-gray}{gray}{0.95}

\begin{document}

\title{Start from Zero: Triple Set Prediction for Automatic Knowledge Graph Completion}

\author{
Wen Zhang, \thanks{Wen Zhang, School of Software Technology, Zhejiang University.}
Yajing Xu, \thanks{Yajing Xu, College of Computer Science of Technology, Zhejiang University.}
Peng Ye, \thanks{Peng Ye, China Mobile (Zhejiang) Innovation Research
Institute Co., Ltd.}
Zhiwei Huang, \thanks{Zhiwei Huang, School of Software Technology, Zhejiang University}
Zezhong Xu, \thanks{Zezhong Xu, College of Computer Science of Technology, Zhejiang University.}
Jiaoyan Chen, \thanks{Jiaoyan Chen, Department of Computer Science, The University of Manchester \& Department of Computer Science, University of Oxford.}
Jeff Z. Pan, \thanks{Jeff Z. Pan, School of Informatics, The University of Edinburgh.}
Huajun Chen \thanks{Huajun Chen, College of Computer Science of Technology, Zhejiang University. Corresponding author.}
}


\markboth{Journal of \LaTeX\ Class Files,~Vol.~14, No.~8, August~2021}%
{Shell \MakeLowercase{\textit{et al.}}: A Sample Article Using IEEEtran.cls for IEEE Journals}

\IEEEpubid{0000--0000/00\$00.00~\copyright~2021 IEEE}

\maketitle

\begin{abstract}
Knowledge graph (KG) completion aims to find out missing triples in a KG. Some tasks, such as link prediction and instance completion, have been proposed for KG completion. They are triple-level tasks with some elements in a missing triple given to predict the missing element of the triple. However, knowing some elements of the missing triple in advance is not always a realistic setting. In this paper, we propose a novel graph-level automatic KG completion task called \textit{Triple Set Prediction (TSP)} which assumes none of the elements in the missing triples is given. TSP is to predict a set of missing triples given a set of known triples. To properly and accurately evaluate this new task, we propose 4 evaluation metrics including 3 classification metrics and 1 ranking metric, considering both the partial-open-world and the closed-world assumptions. Furthermore, to tackle the huge candidate triples for prediction, we propose a novel and efficient subgraph-based method {\model} that can predict the triple set fast. To fairly compare the TSP results, we also propose two types of methods {\baserule} and {KGE-TSP}  applying the existing rule- and embedding-based methods for TSP as baselines. During experiments, we evaluate the proposed methods on two datasets extracted from Wikidata following the relation-similarity partial-open-world assumption proposed by us, and also create a complete family data set to evaluate TSP results following the closed-world assumption. Results prove that the methods can successfully generate a set of missing triples and achieve reasonable scores on the new task, and \model\; performs better than the baselines with significantly shorter prediction time. The datasets and code for experiments are available at https://github.com/zjukg/GPHT-for-TSP.
\end{abstract}


\section{Introduction}
\IEEEPARstart{K}{nowledge} representation and reasoning is one of the key research topics of Artificial Intelligence and has been widely investigated. Knowledge graphs (KG)~\cite{PVGW2017,Pan2017b}, representing facts in the world as triples in the form of \textit{(head entity, relation, tail entity)}, abbreviated as  $(h,r,t)$, is a simple yet effective way for knowledge representation. In recent years, many KGs have been constructed, such as Freebase \cite{bollacker2008freebase}, Wikidata \cite{vrandevcic2014wikidata} and YAGO \cite{pellissier2020yago} for general purpose, and the product KGs \cite{zhang2021billion,dong2018challenges} from Alibaba and Amazon for e-commerce. These KGs are knowledge providers for diverse applications such as searching \cite{PTT2009,rudnik2019searching,GZCL+2019}, question answering \cite{DBLP:conf/naacl/YasunagaRBLL21,HUBP2023},   recommendation \cite{wang2019kgat} and explanations~\cite{CLPHC2018}. Most of these applications rely on triples from KGs, and their quality directly determines to what extent the KGs could contribute to the applications. However, it is widely known that most KGs suffer from incompleteness, making KG completion an important task.

\begin{figure}[t]
    \centering
    \includegraphics[width=0.46\textwidth]{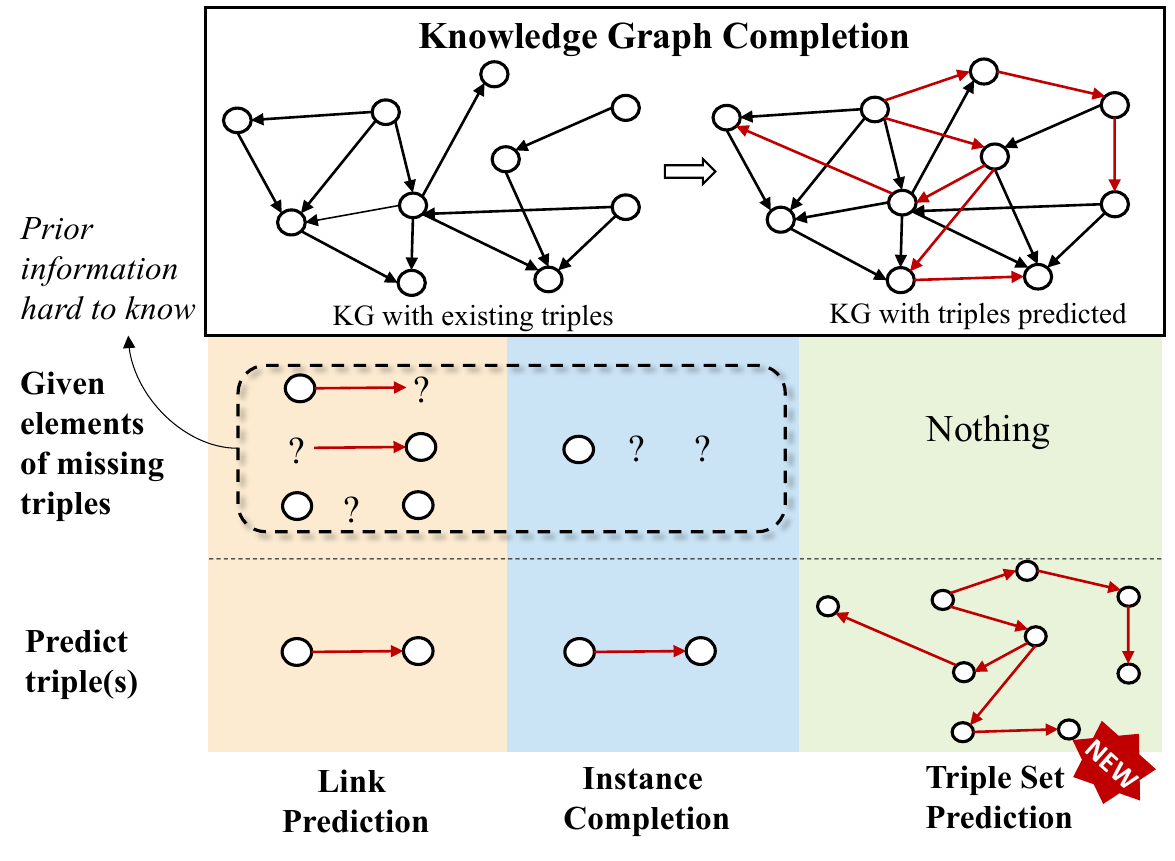}
    \caption{Comparison between the KGC tasks of link prediction and instance completion, and the new KGC task triple set prediction proposed in this paper.}
    \label{fig:intro}
\end{figure}

KG completion (KGC) aims at adding missing but correct triples to KGs, for which many machine learning tasks have been proposed. Among current literature, the triple-level link prediction task is the most widely studied task, targeting predicting the missing element of a triple given the other two elements including tail entity prediction $(h,r,?)$, head entity prediction $(?, r, t)$, and relation prediction $(h, ?, t)$. As pointed out by Rosso et al. \cite{reta}, such a link prediction task is often impractical due to the strong assumption of knowing two elements of a missing triple. Thus instance completion\cite{reta} task, i.e. $(h, ?,?)$, is proposed and studied, which regards a head entity as an instance and completes its semantics by predicting its associated relation-tail pair. Instance completion assumes the list of entities with relation-tail pairs missing are known, which also departs from many real-life KGC use cases where we do not know which entity should be completed. 
For example, given an incomplete e-commerce KG containing many types of entities, such as products, brands, shops, and users, discovering entities with information missing and creating a list of entities to be complete are challenging and require much manual work.

In this paper, with the ultimate goal of KGC in mind, we propose an automatic KGC task called \textit{\underline{t}riple \underline{s}et \underline{p}rediction (TSP)}. 
The task is to output a set of missing triples given a set of known triples in a KG, as shown in Figure \ref{fig:intro}, which exactly matches the goal of KGC.
Specifically, TSP methods are expected to predict all the elements of each missing triple including the head entity, relation, and tail entity, and output a set of missing triples that are believed to be true.
With TSP methods, we could accomplish KGC starting from zero based on the existing triples. 
Thus we believe triple set prediction task is worth to be researched towards automatic KGC.
For the new KGC task, we investigate the following two research questions: (1) how to fairly compare different TSP results, and (2) how to develop efficient and effective TSP methods.

Proposing reasonable evaluation metrics for TSP is challenging. The evaluation metrics should consider both
the size of the predicted triple set and the number of correct triples in the triple set. Specifically, a high-quality predicted triple set is expected to include as many true triples as possible, and as few false triples as possible. 
For example, suppose we have three predicted triple sets, $set_1$ containing 3 triples that are all true, $set_2$ containing 10000 triples with 1000 of them to be true, and $set_3$ containing 1000 triples including 800 true triples. We would expect $set_3$ to be evaluated as better than $set_1$ and $set_2$, since $set_1$ contains too few true triples and $set_2$ contains too many false triples. 
On the other hand, the open-world assumption in knowledge graph representation should also be considered that the truth value of triples not included in the KG are unknown, i.e. could be true or false. 
With the above challenges in mind, we propose 4 evaluation metrics, including 3 classification metrics $JPrecision$,  $STRecall$ and ${F_{TSP}}$, and 1 ranking metric $RS_{TSP}$.

Proposing TSP methods is also challenging. Firstly, for TSP methods, there is no input during prediction. Thus planning the prediction steps and designing the corresponding training stages and targets should be additionally considered when proposing TSP methods. 
Secondly, the number of candidate triples for TSP is large. Theoretically, the candidate number is $ |\mathcal{E}| \times |\mathcal{R}| \times |\mathcal{E}| - |\mathcal{T}|$ where $\mathcal{E},\mathcal{R}, \text{and } \mathcal{T}$ are the set of entities, relations, and triples in a KG, and $|\mathcal{X}|$ is the length of set $\mathcal{X}$. 
With a small toy KG containing $1000$ entities, $100$ relations, and $100000$ triples as an example, the number of missing candidates is $1000 \times 100 \times 1000 - 100000 \approx 10^{8}$.
The number is even larger for real-life KGs containing thousands of entities and more than hundreds of relations. 
With above challenges in mind, we propose to make TSP in two steps. The first step is to predict a set of head-tail entity pairs with relations missing. The second step is to predict the missing relations between each head-tail pair. Our method reduces the candidate space effectively through the first step with \underline{g}raph \underline{p}artition and \underline{h}ead-\underline{t}ail entity pair modeling, thus we name our method as GPHT. Specifically, given a knowledge graph $\mathcal{G}$, we part $\mathcal{G}$ into many distinct subgraphs and regard two entities included in the same subgraph as candidate head-tail entities in the missing triples. Then we train a head-tail entity modeling module in a meta-learning setting to output entity pairs in each subgraph that are likely to miss relations. With predicted head-tail entity pairs, we apply KG embedding methods to predict the missing relations and generate the final predicted triple set.

To fairly compare the TSP results, we propose to adapt rule- and embedding-based KGC methods to TSP task as baselines, including RuleTensor-TSP, HAKE-TSP, and HAKE-TSP. 
For evaluation datasets, we extract two datasets, Wiki79k and Wiki143k from Wikidata with different scales and zero entity overlaps. 
Since Wikidata is incomplete, we evaluate the baselines and the GPHT method under the relation similarity-based partial-open-word assumption that we propose.
We also create a relatively complete dataset CFamily and evaluate the results under the close-world assumption.
During the experiments, we apply to recently proposed effective KGE methods HAKE and PairRE.
The results show that GPHT achieves the best results on two wiki datasets, and comparable results on  CFamily dataset.
More importantly, GPHT has a significant shorter predicting time than baselines, showing the efficiency of GPHT on TSP task. 

In summary, our contributions are
\begin{itemize}
    \item We introduce a new task Triple Set Prediction for automatic KG completion with $4$ evaluation metrics from the classification and ranking perspectives.  
    \item We propose a novel TSP method GPHT and adapt rule- and embedding-based KGC methods to TSP task.
    \item We experimentally prove that GPHT is more effective and efficient for TSP than baselines.  
\end{itemize}

\section{Triple Set Prediction Task}
\label{sec:task_definition}
\subsection{Task Definition}
A KG is $\mathcal{G}=\{ \mathcal{E}, \mathcal{R}, \mathcal{T}\}$. $\mathcal{E}$ is the entity set that includes individuals, such as persons, locations and organizations. $\mathcal{R}$ is the relation set, including relationships between entities, such as \textit{hasFriend} and \textit{locatedIn}. $\mathcal{T} = \{ (h, r, t) | h \in \mathcal{E}, r\in \mathcal{R}, t\in \mathcal{E}\}$ is the triple set, where $h$, $r$, and $t$ are the head entity, relation, and tail entity of the triple. An example of such a triple is \textit{(West Lake, locatedIn, Hangzhou)}.

KGC aims at finding out missing but correct triples for a KG.
There are two general approaches.
One is extracting triples from external resources such as unstructured text and (semi-)structured tables. The other is infer missing triples based on known triples in the KG, which has been attempted to be addressed by tasks such as link prediction and instance completion. The new task TSP belongs to the second approach.

\begin{definition}
(Triple Set Prediction (TSP)) Given a KG $\mathcal{G}$, triple set prediction is to predict a set of missing triples $\mathcal{T}_{predict}$ which are supposed to be true but do not exist in $\mathcal{G}$. For Learning and evaluating TSP models, a training dataset $\mathcal{G}_{train} = \{ \mathcal{E}, \mathcal{R}, \mathcal{T}\}$ is given to learn the model, and a test triple set $\mathcal{T}_{test} = \{ (h,r,t)| h\in \mathcal{E}, r\in \mathcal{R}, t\in \mathcal{E}, (h,r,t) \notin \mathcal{T} \}$ is given to evaluate the model. 
\end{definition}

\begin{figure*}[t]
    \centering
\includegraphics[width=0.8\textwidth]{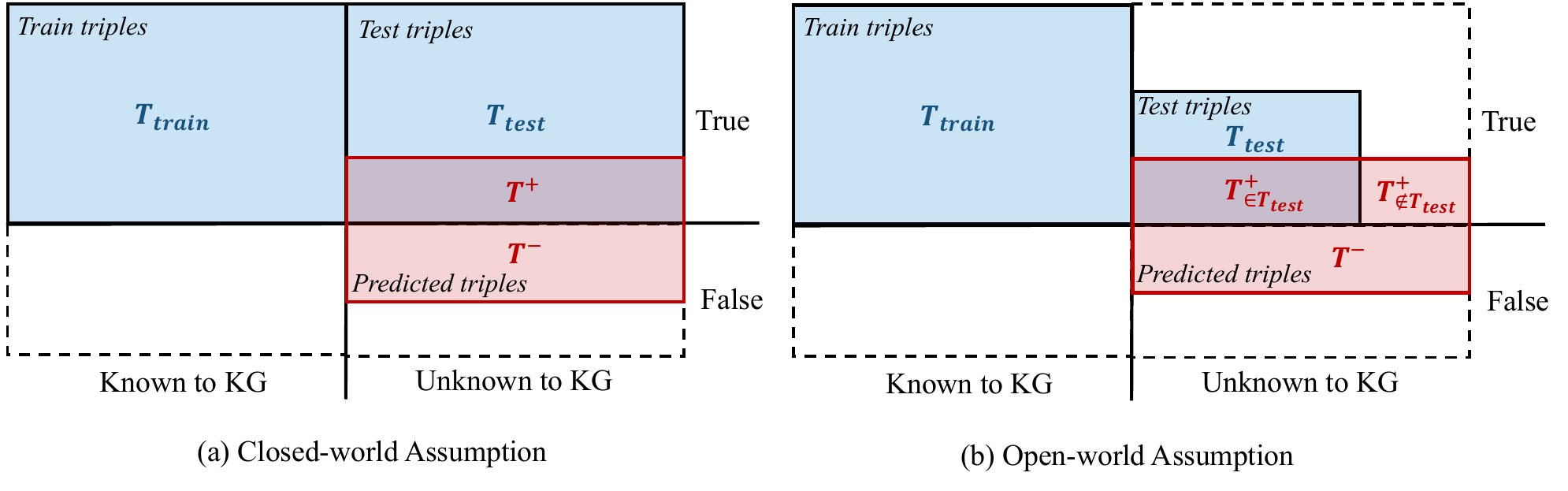}
\vspace{-4mm}
    \caption{Overview of data for triple set prediction under closed-world assumption (a) and open-world assumption (b).}
    \label{fig:data}
\end{figure*}

\subsection{Evaluation Metrics}
\label{sec:evaluation_metrics}
Before introducing the evaluation metrics, we first introduce the world assumptions related to KG representation.   
\subsubsection{Close-World, Open-World, and Relation Similarity-based Partial-Open-World Assumptions}
 According to whether the triple is known to KG and its truth value, there are four types, known-true, known-false, unknown-true, and unknown-false triples. 
In Figure \ref{fig:data}, we show the overview of KG data under closed-world assumption and open-world assumption.

In the Close-World Assumption (CWA), triples not in the KG are supposed to be false. Under the CWA, triples in the train set are known-true and test triples are unknown-true. The predicted triple set $\mathcal{T}_{predict}$ could be divided into two sets, true set $\mathcal{T}^+$ including  triples in $\mathcal{T}_{test}$ and false set $\mathcal{T}^-$ including  triples not in $\mathcal{T}_{test}$, as shown in Figure \ref{fig:data}(a). Thus the target of TSP under the CWA is to maximize the $\mathcal{T}^+$ and minimize $\mathcal{T}^-$. CWA is suitable for KGs that are known to be complete.

While it is known that most KGs are incomplete, thus 
KG representation follows the Open-World Assumption (OWA). With OWA, the truth value of unknown triples is unknown, that is the unknown triples could be either unknown-true or unknown-false. Under OWA, triples in the train set are known-true, and triples in test set are unknown-true. But the test triple set does not include all unknown-true triples. 
Based on the truth value of triples, we could theoretically classify $\mathcal{T}_{predict}$ into three distinct sets, as shown in Figure \ref{fig:data}(b). The first set is unknown-true triples included in $\mathcal{T}_{test}$ denoted as $\mathcal{T}^+_{\in\mathcal{T}_{test}}$. The second set is unknown-true triples that are not included in $\mathcal{T}_{test}$ denoted as $\mathcal{T}^+_{\notin \mathcal{T}_{test}}$. The third set is unknown-false triples denoted as $\mathcal{T}^-$. Formally, $\mathcal{T}_{predict}=\mathcal{T}^+_{\in \mathcal{T}_{test}}\cup\mathcal{T}^+_{\notin \mathcal{T}_{test}}\cup\mathcal{T}^-$. Thus with OWA, the target of TSP could be more clearly expressed as making $\mathcal{T}^+_{\in \mathcal{T}_{test}}\cup\mathcal{T}^+_{\notin \mathcal{T}_{test}}$ as large as possible and making $\mathcal{T}^-$ as small as possible. 

While this is a theoretical classification, it is impossible to distinguish  $\mathcal{T}^+_{\notin \mathcal{T}_{test}}$ and $\mathcal{T}^-$ under OWA. Thus to enable evaluation, we propose the   Relation Similarity-based Partial-Open-World Assumption for a more accurate evaluation of TSP results on incomplete KGs. 

\begin{definition}
    (Relation Similarity-based Partial-Open-World Assumption (RS-POWA) )  
    In a given KG $\mathcal{G}=\{ \mathcal{E}, \mathcal{R}, \mathcal{T}\}$, for each triple $tri = (h,r,t) \notin \mathcal{T}$ with $h \in \mathcal{E}$, $t \in \mathcal{T}$ and $r \in \mathcal{R}$, $tri$ is regarded as false if there exists another relation $r' \in \mathcal{R}$ ($r' \ne r$) such that $(h, r', t) \in \mathcal{T}$ and $r$ is not similar to $r'$ (i.e., $sim(r, r') < \theta$ where $sim$ is a function that calculates the similarity score between two relations, and $\theta$ denotes a given threshold); otherwise, the truth value of $t$ is unknown. 
\end{definition}

In this RS-POWA definition, given two entities $h$ and $t$ in a KG, if one relation $r'$ between them is known, we assume that all relations between them are known. 
Considering that two entities might have two similar relations, such as $hasFather$ and $hasParents$, we regard $(h, r, t)$ as false if  $r'$ is not similar.
We define the similarity between  $r$ and $r'$ as
\begin{equation}
    sim(r, r') = Max(\frac{|\mathcal{P}_{r}\cap\mathcal{P}_{r'}|}{|\mathcal{P}_{r}|}, \frac{|\mathcal{P}_{r}\cap\mathcal{P}_{r'}|}{|\mathcal{P}_{r'}|})
\end{equation}
where $\mathcal{P}_{r}$ is the set of entity pairs $(e_1, e_2)$ that has relation $r$, i.e. $(e_1, r, e_2)\in \{\mathcal{T}_{train}\cup \mathcal{T}_{test}\}$. If the similarity between $r$ and $r'$ is larger than the threshold $\theta=0.8$, we regard them as similar.  
Based on RS-POWA, we can distinguish a set of false triples from $\mathcal{T}_{predict}$.

\subsubsection{Evaluation metrics}
\paragraph{Classification Metrics} 
From the perspective of classification, the goal of TSP is to include more positive and fewer negative triples. 
Thus we first introduce the positive triple set $\mathcal{T}_{predict}^+$ and negative triple set $\mathcal{T}_{predict}^-$ used to evaluate TSP results under CWA and RS-POWA. 

Under CWA, 
\begin{align}
    \mathcal{T}_{predict}^{CWA+} &= \mathcal{T}_{predict} \cap \mathcal{T}_{test} \\
    \mathcal{T}_{predict}^{CWA-} &=  \mathcal{T}_{predict} - \mathcal{T}_{predict}^{CWA+} \\
    \mathcal{T}_{predict}^{CWA} &= \mathcal{T}_{predict}^{CWA+} \cup \mathcal{T}_{predict}^{CWA-} = \mathcal{T}_{predict}
\end{align}

Under RS-POWA,
\begin{align}
\mathcal{T}_{predict}^{POWA+} &= \mathcal{T}_{test} \cap \mathcal{T}_{predict}, \label{equ:t_predicted_powa+}\\
\mathcal{T}_{predict}^{POWA-} &= \{(h,r,t)| (h,r,t)\in \mathcal{T}_{predict}, (h,r,t)\notin \mathcal{T}_{test},\nonumber\\
&\;\;\;\;\; \exists r'\in \mathcal{R}\;\; (h,r',t) \in (\mathcal{T}_{train}  \cap \mathcal{T}_{test}) \\
&\;\;\;\;\; \land sim(r,r')<\theta\}, \\
\mathcal{T}_{predict}^{POWA} &= \mathcal{T}_{predict}^{POWA+} \cup \mathcal{T}_{predict}^{POWA-} \label{equ:t_predicted_powa}
\end{align}
where $\mathcal{T}_{predict}^{WA+}$ and $\mathcal{T}_{predict}^{WA-}$ are the positive and negative triple set that could be recognized in the predicted triple set $\mathcal{T}_{predicted}$ under the assumption 
WA $\in$ \{CWA, RS-POWA\}.
$\mathcal{T}_{predict}^{WA}$ is a set of triples that could be either labeled as positive or negative in the $\mathcal{T}_{predicted}$. When WA = CWA,  $\mathcal{T}_{predict}^{WA} = \mathcal{T}_{predicted}$, and when WA = RS-POWA, $\mathcal{T}_{predict}^{WA} \neq \mathcal{T}_{predicted}$. 
 
Following traditional classification metrics, we propose $3$ metrics, including Joint Precision ($JPrecision$), Squared Test Recall (${STRecall}$), and TSP score ($F_{TSP}$) as follows
\begin{align}
    &JPrecision= \frac{1}{2}(\frac{|\mathcal{T}_{predict}^{WA+}|}{|\mathcal{T}_{predict}^{WA}|} + \frac{|\mathcal{T}_{predict}^{WA+}|}{|\mathcal{T}_{predict}|}), \\
    & STRecall=({\frac{|\mathcal{T}_{predict}^{WA+}|}{|\mathcal{T}_{test}|}})^{\frac{1}{2}}, \\
    &F_{TSP} = \frac{2\times({STRecall} \times JPrecision)}{{STRecall} +  JPrecision}.
\end{align}
$JPrecision$ is the average percentage of unknown-true triples in $\mathcal{T}_{predict}^{WA}$ and $\mathcal{T}_{predict}$.
$STRecall$ is the percentage of predicted test triples.
The square root operation is included in $STRecall$ because of the large number of candidate triples  of TSP.
$JPrecision$ could be trickily made large by intentionally reducing the size of the predicted triple set. For example, outputting one triple that is included in the test set makes $JPrecision$ as 1. 
Similarly, ${STRecall}$ is 1 if the model directly outputs all candidate triples.
While the predicted set should be neither too small nor too large, 
we propose $F_{TSP}$ as a balanced score of $JPrecision$ and ${STRecall}$.

\paragraph{Ranking Metric} Since TSP methods might give each predicted triple a score indicating their truth value, which is usually the larger, the more likely the predicted triple is true.  The classification metrics do not take the score into consideration. Thus we propose a ranking score metric $RS_{TSP}$ to encourage a model not only to predict the triples in the test set as much as possible but also to give them a higher score. 

Given $\mathcal{T}_{test}$ and $\mathcal{T}_{predict}$, we first rank the predicted triples with scores in descent order, resulting in an ordered triple list $\overrightarrow{\mathcal{T}}_{predict}$, 
then we give rank score for the $i$th triple $tri_i \in \overrightarrow{\mathcal{T}}_{predict}$ a score as follows, where $i$ starts from $1$
\begin{equation}
    rs_{tri_i} = \begin{cases}
     \frac{1}{i}, \quad & tri_{i} \in \mathcal{T}^{WA+}_{predict} \\
    -\frac{1}{i}, \quad & tri_{i} \in \mathcal{T}^{WA-}_{predict}
    \end{cases}
\end{equation}
and we calculate the final $RS_{TSP}$ as 
\begin{equation}
    RS_{TSP} = \sum_{trp_{i} \in \mathcal{T}_{predict}^{WA}} rs_{trp_i}
\end{equation}
A larger $RS_{TSP}$ means a better prediction. $RS_{TSP}$ ensures that additionally predicting a true triple makes the ranking score higher and additionally predicting a false triple makes the ranking score lower. There are two theorems. 

\begin{thm}
    Given two predicted triple set $\mathcal{T}_{predict}^1$ and  $\mathcal{T}_{predict}^2$, if $\mathcal{T}_{predict}^2 = \mathcal{T}_{predict}^1 \cup \{(h,r,t)\}$, if $(h,r,t)\in \mathcal{T}_{predict}^{WA+}$, then $RS_{TSP}^2>RS_{TSP}^1$. 
\end{thm}

\begin{thm}
    Given two predicted triple set $\mathcal{T}_{predict}^1$ and  $\mathcal{T}_{predict}^2$, if $\mathcal{T}_{predict}^2 = \mathcal{T}_{predict}^1 \cup \{(h,r,t)\}$, if $(h,r,t)\in \mathcal{T}_{predict}^{WA-}$, then $RS_{TSP}^2<RS_{TSP}^1$. 
\end{thm}

$RS_{TSP}$ also ensures that for two predicted triple sets with the same size, the one ranking positive triples more ahead will get a higher $RS_{TSP}$ score, and there is a theorem as follows:  
\begin{thm}
Given two predicted triple set with the same elements that $\mathcal{T}_{predict}^1 = \mathcal{T}_{predict}^2$ but different order that $\overrightarrow{\mathcal{T}}_{predict}^1 \ne \overrightarrow{\mathcal{T}}_{predict}^2$, if exchange the $i$th triple and the $j$th in $\overrightarrow{\mathcal{T}}_{predict}^1$ results $\overrightarrow{\mathcal{T}}_{predict}^2$ where $i<j$,  
and $tri_i\in\mathcal{T}_{predict}^{WA+}$ and $tri_j\in\mathcal{T}_{predict}^{WA-}$, then $RS_{TSP}^2 < RS_{TSP}^1$.
\end{thm}

\new{
For the correlation between the metrics, there is a trade-off between $STRecall$ and $JPrecision$ that 
increasing one often leads to a decrease in the other. $F_{TSP}$ is a balanced score of $STRecall$ and $JPrecision$, 
and a good $F_{TSP}$ is a necessary but not sufficient condition for a good $RS_{TSP}$.}

\begin{figure}
    \centering
    \includegraphics[width=0.5\textwidth]{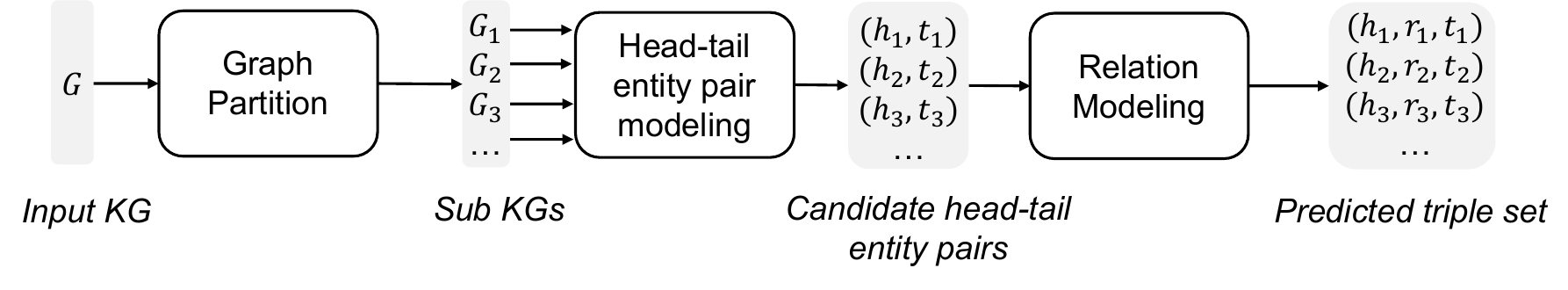}
    \vspace{-5mm}
    \caption{Overview of our method GPHT.}
    \vspace{-5mm}
    \label{fig:method-overview}
\end{figure}

\section{Method}
In this section, we introduce a novel TSP method named \textbf{GPHT}  with \underline{g}raph \underline{p}artition and \underline{h}ead-\underline{t}ail entity pair prediction. The main challenge of TSP is the huge candidate space. Theoretically, the number of candidates for a KG  $\mathcal{G}=\{\mathcal{E}, \mathcal{R}, \mathcal{T}\}$ is $|\mathcal{E}| \times |\mathcal{R}| \times |\mathcal{E}| - |\mathcal{T}|$. The number of candidate triples could reach $10^8$ for a KG with thousands of entities. A good TSP method should effectively reduce the candidate triple space. The effectiveness refers to getting ride of impossible triples at a low cost as much as possible. To reduce the candidate space, we propose GPHT based on the following two assumptions:
\begin{itemize}
    \item (Space assumption) If the length of the shortest path between two entities is large in a KG, 
    it is very likely they have no relationship. For example, in a family relationship KG, if there is no path with a length less than 5 between two persons, these two persons are likely to belong to two distinct families and have no connections. 
    \item (Semantic assumption)  Given two entities with short paths between them in KG, it is also possible they do not have relationships due to semantic mismatching. With a family KG as an example again, a father is less likely to be directly linked to the entity Female, though there might be a path between them with a mother as an intermediate entity. 
\end{itemize}
Thus, in GPHT, we propose to apply graph partition to reduce the candidate space in the space level and apply head-tail entity pair prediction to reduce the candidates in the semantic level.

\subsection{Overview of GPHT}
\label{sec:method-overview}
As shown in Figure \ref{fig:method-overview}, there are three modules in GPHT:

\begin{itemize}
    \item \textbf{Graph partition} is to part $\mathcal{G}$ into several subgraphs containing a comparable number of entities, generating a subgraph set $\mathcal{G}_{part} = \{\mathcal{G}_1, \mathcal{G}_2,..., \mathcal{G}_m\}$. Following the space assumption, we regard entities within a subgraph as candidate entity pairs that have missing relations.
    \item \textbf{Head-tail entity modeling} 
    is to model which entity pairs within a subgraph are likely to have missing relations. Given $\mathcal{G}_{part} = \{\mathcal{G}_1, \mathcal{G}_2,..., \mathcal{G}_m\}$, we regard $\mathcal{G}_i$ as a sample and train a relational graph neural network to get the entity and relation representation in $\mathcal{G}_i$, and train an attention-based network to predict whether there are missing relations between two entities. This step will output a set of head-entity pairs $\mathcal{P}_{ht}=\{(h_1, t_1), (h_2, t_2), ...\}$ that are likely to have missing relations. 
    \item \textbf{Relationship modeling} is to model the missing relationships between two entities. Given $\mathcal{P}_{ht}=\{(h_1, t_1), (h_2, t_2), ...\}$, we train a KG embedding model to predict the relations between each $(h_i, t_i) \in \mathcal{P}_{ht}$. This step will output the final predicted triple set $\mathcal{T}_{predict}$. 
\end{itemize}
Next, in Section \ref{sec:method-gp}, Section \ref{sec:method-htm}, and Section \ref{sec:method-rm}, we introduce the three modules in detail.
Finally, we show how to predict the missing triple set given a KG in Section \ref{sec:method-predicting}.

\subsection{Graph Partition}
\label{sec:method-gp} 

Given $\mathcal{G}$, we part it into several parts and conduct the completion within each part.
After partition, entities included in the same part are regarded as candidate head-tail entity pairs of missing triples. With this step, we could reduce the number of candidate triples from $n_e \times n_r \times n_e$ to approximately $n_{e_G} \times n_r \times n_{e_G} \times m$, where $n_e$ and $n_r$ is the number of entities and relations in $\mathcal{G}$, and $n_{e_G}$ is the average number of entities contained in each subgraph. 
Since $n_{e_G}$ is much smaller than $n_e$ and $n_e$ is equal to $n_{e_G} \times m$, we have $(n_{e_G} \times n_r \times n_{e_G} \times m) \ll (n_e \times n_r \times n_e)$. 

There are two types of graph partition methods ---  vertex-cut partition and edge-cut partition. Vertex-cut partition puts each vertex into one of the subgraphs, but vertex-cut partition will lose the edges between nodes in different subgraphs. Edge-cut partition puts each edge into one of the subgraphs, thus there will be duplicate nodes between subgraphs. In order to maximize the usage of entities and reduce duplication between subgraphs, we propose a ``soft" vertex-cut KG partition method that allows entity overlaps between subgraphs. 
There are two steps in graph partition, primary entity grouping and entity group fine-tuning.
The primary entity grouping step parts the majority of the entities into subgroups and outputs a primary entity group set and an ungrouped entities set.
The entity group fine-tuning puts the ungrouped entities into primary entity groups and outputs the final $\mathcal{G}_{part}$.

\subsubsection{Primary Entity Grouping}
During the partition, we maintain an ungrouped entity set $\mathcal{E}_{U}$ to record the entities that are not included in any entity groups. We initialize it as $\mathcal{E}$. Our goal is to make  $\mathcal{E}_{U}$ empty. 
The overall process is shown in Algorithm \ref{alg1}.

A KG $\mathcal{G} = \{ \mathcal{E}, \mathcal{R}, \mathcal{T}\}$ is not ensured to be a connected graph. Entities in two distinct parts of the KG are not likely to have relations. Thus, we first detect the distinct parts in the KG that $\mathcal{G} = \{\mathcal{G}_1, \mathcal{G}_2, ..., \mathcal{G}_n \}$, where the intersection of entity set $\mathcal{E}_i$ for $\mathcal{G}_i$ and  $\mathcal{E}_j$ for $\mathcal{G}_j$ ($i\ne j$) is empty. 
And for any two entities $e_1 \in \mathcal{E}_i$ and $e_2 \in \mathcal{E}_j$, $\nexists r\in \mathcal{R}$ that $ (e_1, r, e_2)\in \mathcal{G}$ since they are from distinct parts of the KG. Thus $\mathcal{E}$ is divided into  mutually exclusive entities sets that $\mathcal{E} = \{\mathcal{E}_1 \cup \mathcal{E}_2 \cup ... \cup \mathcal{E}_n \}$.

In the entities sets $\{\mathcal{E}_1, \mathcal{E}_2, ... ,\mathcal{E}_n \}$, there might be sets with a small number of entities, which we call \textit{small set}. In order to keep the size of entity groups balanced, we set a $n_{min}$ and a $n_{max}$ to denote the minimum and maximum number of entities that an entity group is expected to have. 
And we merge entity sets with less than $n_{min}$ entities. Specifically, we keep a small set list $\mathbb{E}_{small}$ which is initialized as empty. We traverse $\{\mathcal{E}_1, \mathcal{E}_2, ... ,\mathcal{E}_n \}$. If $|\mathcal{E}_i| < n_{min}$, we merge $\mathcal{E}_i$ with the last entity set in $\mathbb{E}_{small}$ list $\mathcal{E}_{last}$ into a new entity set $\mathcal{E}'$ that $\mathcal{E}' = \mathcal{E}_i \cup \mathcal{E}_{last}$, if $|\mathcal{E}'| < n_{max}$, we add $\mathcal{E}'$ to the entity group set $\mathcal{E}_G$ and remove all $e\in\mathcal{E}'$ from the $\mathcal{E}_{U}$. $\mathcal{E}_G$ is a set used to record the entity groups and is initialized as empty.  

\begin{algorithm}[H]
\SetAlgoLined
\SetKwInOut{Input}{Input}
\SetKwInOut{Output}{Output}

\Input{Subgraph deepth $L$, number of maximum and minimum nodes $n_{max}$ and $n_{min}$, knowledge graph $\mathcal{G}$}
\Output{Entity group list $\mathcal{E}_{G}$, ungrouped entity list $\mathcal{E}_{U}$}
\BlankLine
\emph{Initialize $\mathcal{E}_{G} = \{\;\}, \mathbb{E}_{small} = [ \; ], \mathcal{E}_{U} = \mathcal{E}$ }
\BlankLine
% initialization\; 
Generate distinct connected entity groups  from distinct parts of $\mathcal{G}$, and 
sort them according to number of entities and get $ \{\mathcal{E}_1, \mathcal{E}_2,  ... , \mathcal{E}_n | \forall i,j \; \mathcal{E}_i\cap \mathcal{E}_j = \emptyset, \forall i<j\; |\mathcal{E}_i| \le |\mathcal{E}_j| \}$

\For{i = 1:n}{\label{test}

\If{$|\mathcal{E}_i|<n_{min}$}{

\If{$|\mathcal{E}_i| + |\mathbb{E}_{small}[-1]| < n_{max}$}{
$\mathbb{E}_{small}[-1] \gets \mathcal{E}_i \cup \mathbb{E}_{small}[-1]$ 

\If{$|\mathbb{E}_{small}[-1]|>n_{min}$}{
$\mathcal{E}_G \gets  \mathbb{E}_{small}[-1] \cup \mathcal{E}_G$, remove $\mathbb{E}_{small}[-1]$ from $\mathcal{E}_{small}$, remove $e \in \mathbb{E}_{small}[-1]$ from $\mathcal{E}_U$
}
}
\lElse{add $\mathcal{E}_i$ to $\mathbb{E}_{small}$}
}
}

$\mathcal{E}_{cent} = \mathcal{E}_U$ 

\While{$|\mathcal{E}_{cent}| > 0$}{
Randomly select an $e \in \mathcal{E}_{cent}$ and remove $e$ from $\mathcal{E}_{cent}$

Let $S = \{e\}$

\For{i=1:L}{
get $\mathcal{N}_e^i$ according to Equation (\ref{equ:neighbor})

\If{$i\ne L$}{
$S \gets \mathcal{N}_e^i \cup S $
}
}
\If{$|S|>n_{min}$}{
add $S$ into $\mathcal{E}_G$

remove entity in $S$ from $\mathcal{E}_U$
}

}
\caption{Primary Entity Grouping}
\label{alg1}
\end{algorithm}

Now we have some entities grouped as entity sets in $\mathcal{E}_{G}$ and some entities ungrouped in $\mathcal{E}_{U}$. Our current goal is to make more entities be grouped to generate subgraphs in the later steps. The idea is to iteratively choose an entity from $\mathcal{E}_{U}$ and gather the neighbor entities within $L$-hop as an entity set.
Specifically, in each time of grouping, we randomly choose an entity $e$ from  $\mathcal{E}_{U}$ and extract its neighbor entity set $\mathcal{N}_e$, where we extract the 1-hop, 2-hop, ..., L-hop neighbor entities in order. Each $i$-hop neighbor extraction relies on the $(i-1)$-hop neighbors. Specifically
\begin{align}
    \mathcal{N}_{e}^i &= \bigcup _ {e' \in \mathcal{N}_e^{i-1}} f_n(e', p_i, x) 
    \label{equ:neighbor}
\end{align}
where $f_n(e', p_i, x) $ is to get the 1-hop neighbors of $e'$ with a probability $p_i$.

\begin{equation}
    f_n(e', p, x) = 
    \begin{cases}
    \emptyset \quad  \text{ if } x>p \\
    \{ (e',r,t) | (e',r,t)\in \mathcal{G}, t\in \mathcal{E}_U\} \quad else 
    \end{cases}
\end{equation}
where $x\sim \mathcal{U}(0, 1)$ is a random number following uniform distribution. We noticed that some entity has a huge number of neighbor entities within $L$-hops. Thus to avoid too large entity groups, in the neighbor entity extraction of each hop, we calculate a $p_i$ indicating the percentage of neighbor entity to be included in $\mathcal{N}_{e}^i$ and it is calculated as by 
\begin{equation}
    p_i = 
    \begin{cases}
        1 \quad \text{ if } i = 1 \\
        \sqrt{d_{avg} / (2*\mathcal{N}_{e}^{i-1})} \quad \text{else} 
    \end{cases}
\end{equation}
where $d_{avg}$ is the average degree of all entities in $\mathcal{G}$. $p_i$ relies on the number of entities in $\mathcal{N}_{e}^{i-1}$, and the larger the $\mathcal{N}_{e}^{i-1}$ is, the smaller the $p_i$ is. This is an empirical $p_i$ function that will not lead to too large entity groups.

After getting the neighbor entities of $e$ within $L$-hops, we regard $\mathcal{E}' = \mathcal{N}^{1}_e \cup \mathcal{N}^{2}_e\cup ... \cup \mathcal{N}^L_e \cup \{ e\}$ as an entity group and add $\mathcal{E}'$ to $\mathcal{E}_G$. We remove the entities in $ \mathcal{N}^{1}_e \cup \mathcal{N}^{2}_e\cup ... \cup \mathcal{N}^{L-1}_e \cup \{ e\}$ from $\mathcal{E}_{U}$. 
Please note that entities in  $\mathcal{N}_e^L$ are not excluded from $\mathcal{E}_{U}$, because we think these entities are under-explored that not all of their one-hop neighbor entities are included in $\mathcal{E}^\prime$.  
Thus we allow them to be grouped in multiple entity groups to ensure the information of them in the  KG is fully utilized. 
Finally, we get an updated  $\mathcal{E}_{\mathcal{G}}$ and ungrouped entity set $\mathcal{E}_{U}$ with a small number of entities.

\subsubsection{Entity Group Fine-tuning}

In the entity group fine-tuning step, we add the entities in $\mathcal{E}_U$ into  existing entity groups in $\mathcal{E}_{G}$. 
We firstly randomly sample an entity $e$ from $\mathcal{E}_{U}$, and then traverse the smallest to the largest entity group in $\mathcal{E}_{G}$ until find one entity group $\mathcal{E}'$ that includes $e$. Then we merge the one-hop neighbor entities $\mathcal{N}_e^1$ of $e$ to $\mathcal{E}'$ and remove $e$ from $\mathcal{E}_U$.
This ensures the final entity groups can generate connected graphs and keep the entity group size balanced.  
After iterative entity group fine-tuning, we get an updated entity group list $\mathcal{E}_G$ and  $\mathcal{E}_U = \emptyset$. Algorithm \ref{alg2} shows the overall process.

\begin{algorithm}[H]
\SetAlgoLined
\SetKwInOut{Input}{Input}
\SetKwInOut{Output}{Output}
\Input{Entity group list $\mathcal{E}_{G}$, ungrouped entity list $\mathcal{E}_U$}
\Output{Updated entity group list $\mathcal{E}_{G}$}
\While{$|\mathcal{E}_U| > 0$}{
Randomly choose an entity $e$ from $\mathcal{E}_U$ 

Decently sort entity groups in $\mathcal{E}_{G}$ according to the group size and get $ \overrightarrow{\mathcal{E}_{G}}$ 

\For{$\mathcal{E}' \in \overrightarrow{\mathcal{E}_{G}}$}{
\If{$e \in \mathcal{E}'$}{
update $\mathcal{E}' \gets \mathcal{E}'\cup \mathcal{N}_e^i$

remove $e$ in $\mathcal{E}'$ from $\mathcal{E}_U$ 
}

}

% \For{$([\mathcal{N}], S) \in sort(\mathcal{E}_{G})$}{
% \If{$e$ is connected to $S$}{
% \For{$(e, r, t) \in \mathcal{T}$}{
% add $t$ to $S$
% }
% remove $e$ from $\mathcal{E}_U$
% }
% }
}

\caption{Entity Group Fine-tuning}
\label{alg2}
\end{algorithm}

\subsubsection{ Subgraph Construction} 
For each $\mathcal{E}' \in \mathcal{E}_G$, we construct the subgraph $\mathcal{G}_{\mathcal{E'}}$ by adding all triples with head and tail entity in $\mathcal{E}'$, that 
$\mathcal{G}_{\mathcal{E'}} = \{\mathcal{E}', \mathcal{R}, \mathcal{T}'\}$ where $\mathcal{T}' = \{ (h,r,t) | (h,r,t) \in \mathcal{G}, h\in \mathcal{E}', t \in \mathcal{E}' \}$. After that, we will get a subgraph set $\mathcal{G}_{part} = \{\mathcal{G}_1, \mathcal{G}_2,..., \mathcal{G}_m\}$, where $m$ is the number of entity groups in $\mathcal{E}_G$.

\subsection{Head-tail Entity Modeling (HTEM)}
The HTEM module is to predict head-tail entity pairs in one subgraph that have relations missing. 
Thus the function of HTEM module is $\mathcal{P}_{ht} = HTEM(\mathcal{G}_i)$, where $\mathcal{G}_i \in \mathcal{G}_{part}$, and $\mathcal{P}_{ht} = \{ (h_1, t_1), (h_2, t_2), ...\}$ is a set of predicted candidate head-tail entity pairs.

\begin{figure*}
    \centering
    \includegraphics[width=0.9\textwidth]{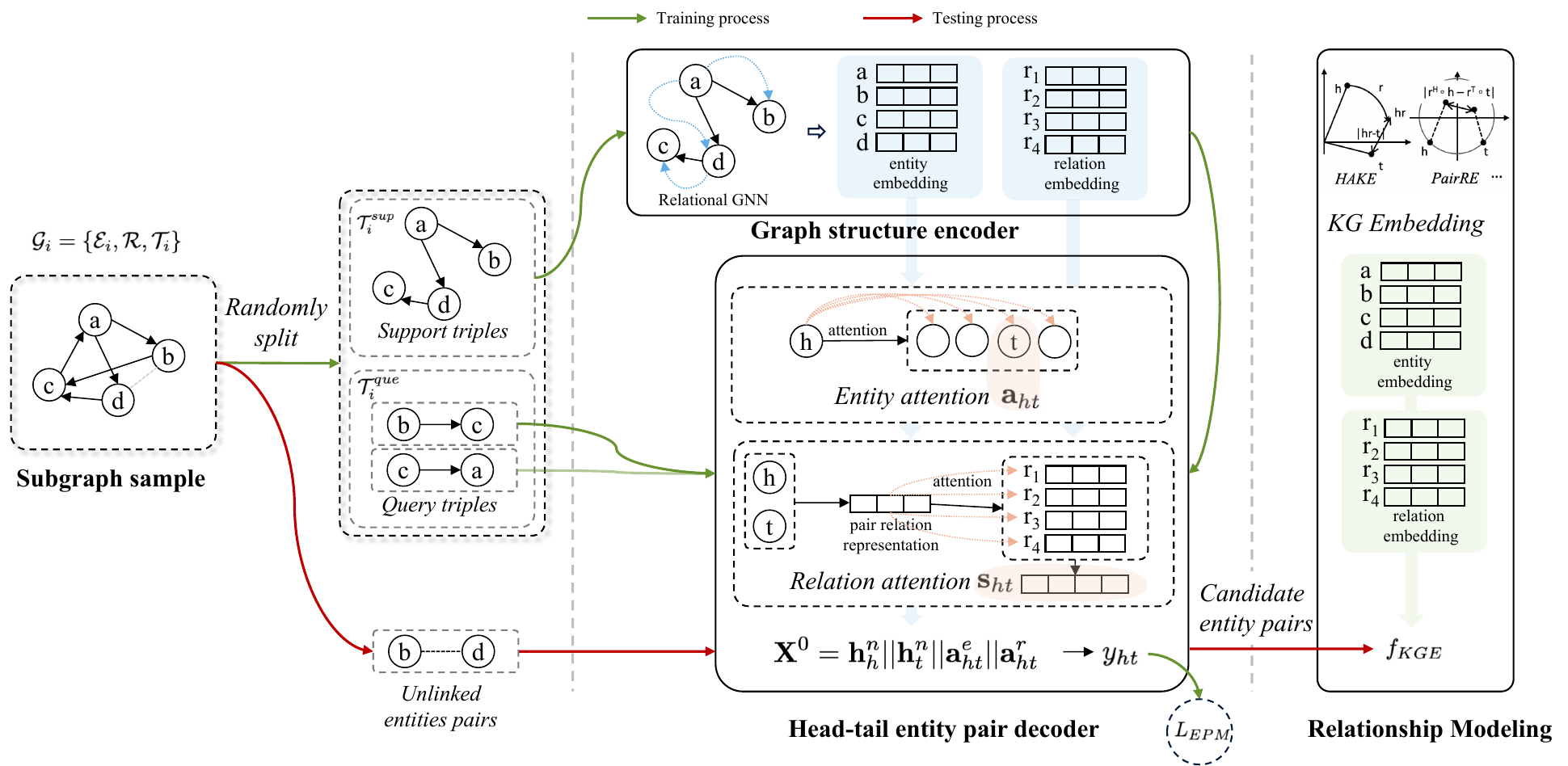}
    \vspace{-3mm}
    \caption{Overview of the training and testing process of GPHT given one subgraph sample.}
    \label{fig:model-specific}
\end{figure*}

To train HTEM module, we follow a meta-learning setting, regarding each subgraph as a sample, enabling HTEM to learn how to predict the head-tail entity pair given a subgraph.  
Specifically, given $\mathcal{G}_i \in \mathcal{G}_{part}$, we randomly split triples into two sets $\mathcal{T}_{i} = \{\mathcal{T}_{i}^{sup}, \mathcal{T}_i^{que}\}$, where $\mathcal{T}_{i}^{sup}$ is the support triple set used to get the representation of entities and relations in $\mathcal{G}_i$, and 
$\mathcal{T}_i^{que}$ is the query set used to calculate the model loss as training objective.
To encode the $\mathcal{T}_{i}^{sup}$, we apply a relational graph neural network (RGNN) that is aware of the relations between entities and could output the structure-awared representation for entities and relations. 
To predict the likelihood of the head-tail entity pairs in $\mathcal{T}_{i}^{sup}$, we design a head-tail entity pair encoder with entity attention and relation attention. The overall process is shown in Figure \ref{fig:model-specific}. 

\subsubsection{Graph Structure Encoder}
We apply a widely used and effective  RGNN,  CompGCN \cite{compgcn}, to capture the graph structure in the subgraph.
Given the $\mathcal{T}_i^{sup}$ of the subgraph sample $\mathcal{G}_i$,
we add the inverse triple $(t, r_{inv}, h)$ of each $(h,r,t) \in \mathcal{G}_i$ into $\mathcal{T}_i^{sup}$, and for each entity $e\in\mathcal{E}_i$, we add a selfloop triple $(e, r_{selfloop}, e)$ to $\mathcal{T}_i^{sup}$.
CompGCN updates the representations of entities by aggregating neighbor information in each layer. 
In the $k$-th layer, entity $e$'s representation $\mathbf{h}_{e}^{k}$ and relation $r$'s representation $\mathbf{h}_r^k$ is updated through 
\begin{equation}
\vspace{-3mm}
\mathbf{h}_e^{k} = f(\sum_{(e, r, e')\in\mathcal{G}} \mathbf{W}_{dir(r)}^k \phi (\mathbf{h}_{e'}^{k-1}, \mathbf{h}_{r}^{k-1})), \; \mathbf{h}_r^{k} =  \mathbf{W}_{rel}\mathbf{h}_{r}^{k-1} \nonumber
\end{equation}
where $\phi()$ is the composition function, and  $\mathbf{W}_{dir(r)} \in \mathbb{R}^{d\times d}$ depends on the direction of the triple that
\begin{equation}
    \mathbf{W}_{dir(r)} = \begin{cases}
\mathbf{W}_{O}, r\in \mathcal{R}\\
\mathbf{W}_{I}, r\in \mathcal{R}_{inv},  \mathcal{R}_{inv} = \{r_{inv}|r\in\mathcal{R}\}\\
\mathbf{W}_{S}, r=r_{selfloop} 
\end{cases} 
\end{equation}
In order to avoid too many parameters for relations, $\mathbf{h}_r^0 = \sum_{i=1}^b \alpha_{i,r}\mathbf{v}_{i}$ is initialized through $b$ base embeddings where $\alpha_{i,r}$ is the learnable weight for the $i$-th base embedding for $r$. Suppose there are $n$ CompGCN layers, the representations from the $n$-th layer are regarded as the structure-awared representation for entities and relations, denoted as $\mathbf{h}_e^n$ for each entity $e$ and $\mathbf{h}_r^n$ for each relation $r$.

\subsubsection{Head-tail Entity Pair Decoder}
\label{sec:method-htm}
This module is used to model the possibility of head-tail entity pairs missing relations, in which we designed an entity attention and a relation attention to enable the module aware of the tail entities and the missing relations between the entity pairs. 

Specifically, given the query triple set $\mathcal{T}_i^{que}$ of $\mathcal{G}_i$, for each head-tail entity pair in the triple $(h,r,t) \in \mathcal{T}_i^{que}$, we calculate the likelihood score $y_{ht}$ through
\begin{align}
    &\mathbf{X}^0 = \mathbf{h}_h^n || \mathbf{h}_t^n || \mathbf{a}_{ht}^e || \mathbf{a}_{ht} ^r\nonumber\\
    &\mathbf{X}^{i+1} = Dropout(LeakyReLU(\mathbf{X}^i(\mathbf{W}^i))), i\in [1, k] \nonumber\\
    &y_{ht} = \sigma(\mathbf{X}^k) \label{eq:ht_score}
\end{align}
where $\mathbf{x}||\mathbf{y}$ means concatenation of two embeddings $\mathbf{x}$ and $\mathbf{y}$. 
$\mathbf{a}_{ht}^e \in \mathbb{R}$ is the entity attention score capturing the relatedness of two entities based on their representation. $\mathbf{a}_{ht}^r \in \mathbb{R}^{n_r}$ is the relation attention vector indicating the potential relation that might be missing between two entity pairs. $n_r$ is the number of relations.
$\mathbf{X}^0$ is input into a $k$ nonlinear layers to get the predicted likelyhood score $y_{ht}$. Finally, the larger $y_{ht}$ is, the more possible relationships between $h$ and $t$ are missing.

\textit{Entity Attention Score.}
$\mathbf{a}_{ht}$ is calculated through 
\begin{align}
    & \mathbf{Q}_h =\mathbf{h}_h^n\mathbf{W}^Q, \quad \mathbf{K}_t = \mathbf{h}_t^n\mathbf{W}^K, \quad {att}(h,t) = \frac{\mathbf{Q}_h\mathbf{K}_t^{\top}}{\sqrt{d}} \\
    & \mathbf{a}_{ht} =\frac{exp(att(h,t))}{\sum_{e \in \mathcal{E}_i} exp( att(h,e) )}, 
\end{align}
$\mathbf{a}_{ht}$ indicates the likelihood between $h$ and $t$. 

\textit{Relation Attention Vector.} 
The intuition of the relation attention vector is that 
if $r$ is predicted to be missing between $h$ and $t$, then it is more likely that there are missing relations between $h$ and $t$. 
There's a lot of research about KG embedding (KGE) \cite{rotate, paire,transe,transh} that could predict the missing relations between entities. Thus we resort to KGEs for $\mathbf{s}_{ht}$ calculation.
In most KGEs, we could derive a mapping function $f_{kge}^r: \mathcal{E} \times \mathcal{E} \mapsto \mathcal{R}$ that could map the head entity embedding $\mathbf{h}$ and the tail entity embedding  $\mathbf{t}$ to a relation based on their space assumptions. 
Thus we calculate the relation attention score as 
\begin{align}
    \mathbf{r}_{ht} = f_{kge}^r(\mathbf{h}_h^n, \mathbf{h}_t^n), \quad \mathbf{s}_{ht}^{(i)} = sim(\mathbf{r}_{ht}, \mathbf{r}_i)
\end{align}
where $\mathbf{s}_{ht}^{(i)}$ is the $i$-th value of the $\mathbf{s}_{ht} \in \mathbb{R}^{n_r}$. $sim(x,y)$ is a similarity function. 
In this work, we applied two recently proposed KGEs, HAKE \cite{hake} and PairRE \cite{paire}.

\textit{For HAKE.} 
HAKE \cite{hake} is a hierarchy-aware KGE model consisting of two parts -- a modulus part and a phase part used to model entities in two different categories. And the score function of HAKE is 
\begin{equation}
    f_{hake}(\mathbf{h}, \mathbf{r}, \mathbf{t}) = - || \mathbf{h}^m \circ \mathbf{r}^m - \mathbf{t}^m || _2 - \lambda||sin((\mathbf{h}^p + \mathbf{r}^p - \mathbf{t}^p)/2)||_1 
    \label{func:hake-function}
\end{equation}
where $\mathbf{x}^m$ and $\mathbf{x}^p$ denotes the modulus and phase representation of the element $x$. 
Thus we calculate the $\mathbf{s}_{ht}^{(i)}$ as 
\begin{align}
& \mathbf{s}_{ht}^{(i)} = \mathbf{r}_{ht}^m \cdot \mathbf{r}_{i}^{mm} + \mathbf{r}_{ht}^p \cdot \mathbf{r}_{i}^p
\end{align}
where $\mathbf{r}_{ht}^m$ and $\mathbf{r}_{ht}^p$ is the modulus representation and phase representation of the inferred relation between $h$ and $t$. $ \mathbf{r}_{i}^{mm}$ and $ \mathbf{r}_{i}^{p}$ is the modulus representation and phase representation of the $i$th relation $r_i$ used for similarity calculation. These representations are calculated as follows
\begin{align}
\mathbf{r}_{ht}^m = \mathbf{t}^m/\mathbf{h}^m,  
\quad \mathbf{r}_{ht}^p = \mathbf{t}^p - \mathbf{h}^p, 
\quad \mathbf{r}_{i}^{mm} = \frac{\mathbf{r}_{i}^{m} + \mathbf{r}_{i}^{b}}{1-\mathbf{r}_{i}^b}
\end{align} 
where $\mathbf{r}_{i}^b$ is the bias representation for relation $r$. 
We split the modulus and phase embedding of element $h$ and $t$ from its representation from the last (the $n$th)  layer of the graph structure encoder that ($e \in \{ h, t\}$)
\begin{align}
 (\mathbf{e}^m || \mathbf{e}^p)  =  \mathbf{W}^e\mathbf{h}_e^n 
 \quad(\mathbf{r}_i^p ||\mathbf{r}_i^m ||\mathbf{r}_i^b) = \mathbf{h}_r^n
\end{align}

\textit{For PairRE.}
PairRE \cite{paire} is a KGE model simultaneously encoding complex relations and multiple relation patterns. 
It uses two vectors for relation representation. These vectors project the corresponding head and tail entities to Euclidean space. 
% where the distance between the projected vectors is minimized.
And the score function of PairRE is
\begin{equation}
    f_{pairre}(\mathbf{h}, \mathbf{r}, \mathbf{t})=-||\mathbf{h}\circ\mathbf{r}^H - \mathbf{t}\circ\mathbf{r}^T||
    \label{func:pairre-function}
\end{equation}
where $\mathbf{h}, \mathbf{r}^H, \mathbf{r}^T, \mathbf{t} \in \mathbb{R}^d$ and $||\mathbf{h}||^2 = ||\mathbf{t}||^2 = 1$.
Based on the assumption of PairRE, we calculate the $\mathbf{s}_{ht}^{(i)}$ as
\begin{align}
\mathbf{s}_{ht}^{(i)} = \mathbf{h} \cdot \mathbf{r}^{H} - \mathbf{t} \cdot \mathbf{r}^{T} 
\end{align}
We get the embedding of elements $h$ and $t$ from 
\begin{align}
\mathbf{e} =  \mathbf{W}^{e}\mathbf{h}_e^{n} 
 \quad(\mathbf{r}^{h} ||\mathbf{r}^{t}) = \mathbf{h}_r^n
\end{align}
\label{equ:pairre}

During training, we define the positive pair set as $\mathcal{T}_{i}^+ = \mathcal{T}_{i}^{que}$, and negative pair set as  $\mathcal{T}_{i}^- = \{(h,r,t)|h\in \mathcal{E}_i, t \in \mathcal{E}_i, r\in \mathcal{R}, (h,r,t)\notin\mathcal{T}_i\}$. 
For positive pairs, we encourage $y_{ht}$ to be large and small for negative pairs. The loss function is 
\begin{align}
    L =  & \sum_{\mathcal{G}_i\in \mathcal{G}_{part}} L_{\mathcal{T}_i} \\
    L_{\mathcal{T}_i} = & \frac{1}{|\mathcal{T}_i^{+}|}
    \sum_{(h,r,t)\in \mathcal{T}^{+}_i} 
    (1-y_{ht}) 
    + \frac{1}{|\mathcal{T}_i^-|}\sum_{(h,r,t)\notin \mathcal{T}_i^-} y_{ht} \nonumber \\
    & + \sum_{(h,r,t)\in \mathcal{T}_i^{sup}}log\sigma(f_{kge}(\mathbf{h}_h^n,\mathbf{h}_r^n,\mathbf{h}_t^n))
    \label{eq:loss}
\end{align}
where $f_{kge}$ is the score function of the KGE, i.e. Equation (\ref{func:hake-function}) for HAKE and Equation (\ref{func:pairre-function}) for PairRE. 
The score loss from KGE is included to  
make the model aware of the truth value of existing triples. This ensures the HTEM module does not destroy what has been learned from the graph structure encoder.

\subsection{Relationship Modeling}
\label{sec:method-rm}
This module is to predict the relationship between head-tail entity pairs. 
As introduced before,
many KGEs could effectively evaluate the  truth value with a score function 
$f_{kge}(\mathbf{h}, \mathbf{r}, \mathbf{t})$ to evaluate the truth value of a triple $(h,r,t)$.
Usually, the scores of true triples are expected to be large and false triples to be small. 
In this work, we make the KGE in relationship modeling the same as KGE applied in the head-tail entity pair decoder, i.e.  
HAKE \cite{hake} and PairRE \cite{paire}. 
We train them on the whole knowledge graph $\mathcal{G}$ with the self-adversarial training loss $L_{kge}$ that 
\begin{align}
L_{kge} =  - & \sum_{(h,r,t)\in \mathcal{G}}( log \sigma(f(\mathbf{h}, \mathbf{r}, \mathbf{t})) \nonumber\\
& - \sum_{i=1}^k p(h_i',r,t_i')log \sigma (-f_{kge}(\mathbf{h}_i', \mathbf{r}, \mathbf{t}_i')) \label{eq:HAKE_loss}
\end{align}
where $\sigma()$ is the sigmoid function. $(h', r, t')$ is the negative triple of $(h,r,t)$ by randomly replacing $h$ or $t$ with other entities in the KG. $p(h_j', r, t_j')$ is got from a weighted softmax function with a hyperparameter $\alpha$ that 
\begin{align}
p(h_i', r, t_i') = \frac{\text{exp } \alpha f_{kge}(\mathbf{e}_{h'}, \mathbf{e}_r, \mathbf{e}_{t'})}{\sum_{i} \text{exp } \alpha f_{kge}(\mathbf{e}_{h_i},\mathbf{e}_r, \mathbf{e}_{t_i})}
\label{eq:HAKE_softmax}
\end{align}

\subsection{Predicting}
\label{sec:method-predicting}
In this section, we introduce how to output the final predicted triple set based on these three modules in three steps. 

Firstly, we get the parted subgraph set $\mathcal{G}_{part} = \{ \mathcal{G}_1, \mathcal{G}_2, ..., \mathcal{G}_m\}$ based on the graph partition module.

Secondly, for each subgraph $\mathcal{G}_i = \{ \mathcal{E}_i, \mathcal{R}, \mathcal{T}_i\}$, we predict the candidate head-tail entity pair based on the well trained HTEM module. Specifically, we regard all unconnected entity pairs as candidate pairs. 
Then, we calculate the pair score of each candidate pair following Equation (\ref{eq:ht_score}) and select the pairs with $y_{ht}$ larger than threshold $\theta_{ht}$ where $\theta_{ht}$ is a hyperparameter. Thus the predicted pair set over $\mathcal{G}_i$ is $\mathcal{P}_{i} = \{(h,t)| h\in\mathcal{E}_i, t \in \mathcal{E}_i, y_{ht}>\theta_{ht}, \nexists r \; (h, r, t)\in \mathcal{T}_i  \}$. Finally, the predicted pair set over $\mathcal{G}_{part}$ is $\mathcal{P}_{ht} = \bigcup _{i=1} ^m\mathcal{P}_i$.

Thirdly, we predict the 
the missing relationships over $\mathcal{P}_{ht}$.
For each pair $(h, t) \in \mathcal{P}_{ht}$, we regard all the relations as candidate relations between them, thus the candidate set is $ \mathcal{C}_{hrt} = \{(h, r_i, t) | (h,t) \in \mathcal{P}_{ht}, r_i\in \mathcal{R} \} $. We calculate the scores of each candidate triple $(h, r_i, t)$ according to the score function of the corresponding KGE. Then we normalize the triple scores through a softmax function that 
\begin{equation}
    s_{hrt}(h,r,t) = exp \frac{f_{kge}(\mathbf{h}, \mathbf{r}, \mathbf{t})}{\sum_{(h,r,t) \in \mathcal{C}_{hrt}} exp f_{kge}(\mathbf{h}, \mathbf{r}, \mathbf{t})}
    \label{equ:kge_softmax}
\end{equation}
Finally, we include triples with score larger than a threshold $\theta_{hrt}$ in the final predicted triples set $\mathcal{T}_{predict}$, i.e. 
\begin{align}
    \mathcal{T}_{predict} = \{ (h,r,t) | (h,r,t) \in \mathcal{C}_{hrt}, s_{hrt}(h,r,t)>\frac{\theta_{hrt}}{| \mathcal{C}_{hrt}|}\} \nonumber
\end{align}
We divide the $\theta_{hrt}$ by the total number of triples in $\mathcal{C}_{hrt}$ to make the hyperparameter $\theta_{hrt}$ setting insensitive to the number of candidates.  
After the softmax function, the larger the number of candidate triples is, the smaller the score is.

\section{Experiment}
\subsection{Datasets}
We construct 3 new datasets for evaluation.
Firstly, we extract two subsets from Wikidata, called \textit{Wiki79k} and \textit{Wiki143k} with zero entity overlap, and test the TSP results on them under the RS-POWA, since Wikidata is incomplete. 
We intentionally make \textit{Wiki143k} significantly larger than \textit{Wiki79k} to explore the impact of dataset size. 
We also construct a relatively complete dataset called CFamily. 
Specifically, based on an initial set of triples about the family relationships between people, we add missing triples following family relationship rules, such as $fatherOf \gets husbandOf \land motherOf$\footnote{Equal to path rule written as $fatherOf(X, Y) \gets husbandOf(X,Z) \land motherOf(Z, Y)$}. CFamily enables us to evaluate the TSP results under the CWA.
For each dataset, we split the triples into train, valid, and test sets. The statistics of the datasets are shown in Table  \ref{tab:datasets}.

\subsection{Baseline methods}
As a new task,
there is no existing method that can be directly applied to TSP. In order to fairly compare to GPHT, we adapt the rule-based and embedding-based methods for link prediction to TSP, named RuleTensor-TSP and KGE-TSP.

\begin{table}[]
    \centering
    \caption{Statistics of datasets in experiments.}
    \resizebox{0.49\textwidth}{!}{
    \begin{tabular}{l c c c c c c c}
    \toprule
        \textbf{Datasets} & \textbf{\#Ent} & \textbf{\#Rel} & \textbf{\#Triple} & \textbf{\#Train} & \textbf{\#valid} & \textbf{\#Test} & \textbf{ Assumption} \\
        \midrule 
        Wiki79k	&7983	&85	&79213 & 57033& 6337 &15843	&RS-POWA \\
        Wiki143k &13928	&109	&143632 &103415 &14190 &28727 &RS-POWA \\
         CFamily	&2378	&12	&22986 &16549 & 1839 &4598 &CWA \\
         % CFamily & 2378 & 12 & 22986 & CWA \& POWA\\
         % WN18RR & 41105 & 11 & 93003 & POWA\\
         % FB15k-237 & 14541 & 237 & 310116 & POWA\\
         \bottomrule
    \end{tabular}
    }
    \label{tab:datasets}
\end{table}

\subsubsection{RuleTensor-TSP} 
We use tensor calculation to simulate the rule reasoning inspired by TensorLog\cite{tensorlog}, thus we name the proposed baseline RuleTensor-TSP. 

In RuleTensor-TSP, the first step is rule mining, including candidate rule sampling and high-quality rule selection. Given a KG $\mathcal{G}$, 
we first add the inverse triple of each triple to the $\mathcal{G}$, i.e., $\mathcal{G} \gets \mathcal{G} \cup \{ (t, r^{-1}, h)| (h,r,t)\in \mathcal{G}\}$. 
We use multiple times of random walks to sample candidate path rules, resulting in a set $\mathcal{R}ule_c$. 
Specifically, we randomly select one entity $e_0$ as the start node and randomly walk with maximum $L$ steps on the $\mathcal{G}$, and repeat this procedure.
In the $i$-th step, we randomly select one relation $r_i$ from triples with entity $e_{i-1}$ as the head entity, and randomly select a tail entity from triples with $e_{i-1}$ and $r_i$ as head entity and relation, where $e_i$ is the selected entity at the $i$th step. 
If there is triple $(e_0, r, e_i) \in \mathcal{G}$, we terminate the random walk and generate a rule  $r \gets r_1\land r_2 \land ... \land r_i$, where $r$ is called the rule head and $r_1\land r_2 \land ... \land r_i$ is the rule body, and add it into $\mathcal{R}ule_c$. 
If there is triple $(e_i, r, e_0) \in \mathcal{G}$, we terminate the random walk and generate a rule $r \gets r_i^{-1}\land r_{i-1}^{-1} \land ... \land r_1^{-1}$ and add it into $\mathcal{R}ule_c$.
If no triples include $e_0$ and $e_i$, we conduct the next step of the random walk. 
In this way, we can generate rules with rule body size no longer than $L$. 
We use two commonly used quality metrics \textit{confidence} ($conf$), and \textit{head coverage} ($hc$)\cite{amie3} to select high-quality rules from $\mathcal{R}ule_c$. To speed up the prediction, we calculate the metrics with tensors. 
We represent each relation $r$ as matrix $\mathbf{M}^r \in \mathbb{R}^{n_e\times n_e }$, where $\mathbf{M}^r_{ij}=1$ if $(e_i, r, e_j) \in \mathcal{G}$, otherwise $\mathbf{M}^{r}_{ij}=0$, and $n_e$ is the number of entities in the KG. With tensor representation, $support$, $conf$ and $hc$ of $rule: r \gets r_1\land r_2 \land ... \land r_k$ is calculated through
\begin{align}
sup(rule) &= \sum\left(\mathbf{M}^r \circ \mathbf{M}^{body}\right), \;\;\mathbf{M}^{body}= f_{[0,1]}\left(\prod_{j=1}^{k} \mathbf{M}^{r_j}\right) \label{eq:rule_body} \nonumber \\
conf(rule) &=\frac{sup(rule)}{\oplus{\mathbf{M}}^{body}}, \; hc(rule) = \frac{sup(rule)}{\oplus{\mathbf{M}}^{r}}
\end{align}
where $f_{[0,1]}(\mathbf{M})$ makes the values larger than $0$ in $\mathbf{M}$ to $1$,
since matrix multiplication could result matrix with $\mathbf{M}'_{ij}>1$ if there are multiple paths could infers $(e_i, r, e_j)$. Specifically, $f_{[0,1]}(\mathbf{M}_{ij}) = 1$ if $\mathbf{M}_{ij} > 1$. 
$\oplus\mathbf{M}$ is the summation of all values in $\mathbf{M}$. $\circ$ is the Hadamard product. Finally, we collect a set of high-quality rules $\mathcal{R}ule_q$ with confidence and head coverage higher than the threshold $\theta_{conf}$ and $\theta_{hc}$.  

The second step is predicting a triple set with multiple iterations of rule inference based on $\mathcal{R}ule_q$.
At the $(t+1)$th iteration, for each rule $rule$ with $r$ as head relation, we calculate the head relation matrix  $(\mathbf{M}^{r})^{t+1}$ including the inferred triples and the existing triples with $r$ as the relation:
\begin{equation}
    (\mathbf{M}^{r})^{t+1} = conf(rule) \times ((\mathbf{M}^{body})^t - f_{[0,1]}(\mathbf{M}^r)^t) + (\mathbf{M}^r)^t
\end{equation}
where $(\mathbf{M}^{body})^t = \prod_{j=1}^{k} (\mathbf{M}^{r_j})^{t}$ and $(\mathbf{M}^r)^0 = \mathbf{M}^r$. If a new $(e_i, r, e_j)$ triple is inferred, i.e., $(\mathbf{M}^{body})^t_{ij}\ne 0 $ and $(\mathbf{M}^r)^t_{ij}=0$, its truth value is marked as the confidence of $rule$.  $(\mathbf{M}^r)^{t+1} = (\mathbf{M}^r)^t$ means all triples that can be inferred by the $rule$ are added into the KG, and we will terminate the inference.  
Suppose we terminate the iteration after $t$ times of iteration, the predicted triple set is 
\begin{equation}
    \mathcal{T}_{predict} = \{((e_i, r, e_j)| (\mathbf{M}^{r})^{t}_{ij}-(\mathbf{M}^r)^{0}_{ij} \ne 0, r\in\mathcal{R} \}
\end{equation}

\subsubsection{KGE-TSP}
Given a KG $\mathcal{G}$, we train the KGE model with triples in $\mathcal{G}$ and get the well-trained entity and relation embeddings and score function $f_{kge}()$. Then we traverse all possible triples $\mathcal{T}_{all} = \{(h,r,t)| h\in \mathcal{E}, r\in\mathcal{R}, t \in \mathcal{E}\}$ and select triples as follows
\begin{equation}
\mathcal{T}_{predict} = \{ (h,r,t)| s_{hrt}(h,r,t)>\frac{\theta_{kge}}{|\mathcal{T}_{all}|}, (h,r,t)\notin \mathcal{T}_{train}\}
\end{equation}
where the calculation of $s_{hrt}$ is the same as Equation (\ref{equ:kge_softmax}).  Note that the number of candidate triples $n_e \times n_r\times n_e$  is large. 
Storing the score of all triples takes a lot of memory. 
Thus we traverse all triples two times to calculate the softmax score without storing the scores. In the first time traversal, 
we initialize a $s=0$, calculate the score of each triple $f_{kge}(\mathbf{h},\mathbf{r},\mathbf{t})$ and add the score to $s$, to simulate $\sum_{(h,r,t)\in \mathcal{T}_{all}} f_{kge}(\mathbf{h},\mathbf{r},\mathbf{t})$. In the second time, we calculate the score of each triple again and divide them by $s$. During the experiment, we select HAKE and PairRE, named HAKE-TSP and PairRE-TSP. 

\subsection{Experiment details}
\textbf{For \baserule}, 
we select the rule length $K$ from $\{3, 4\}$, $\theta_{conf}$ from $\{ 0.98,0.9, 0.85, 0.6, 0.45\}$, and  $\theta_{hc}$ from $\{ 0.05, 0.2, 0.4, 0.65, 0.85\}$. Finally, we got the best results on $\theta_{conf}=0.85, \theta_{hc}=0.05.$ The inference stops if the predicted triple is $20\%$ smaller than the last iteration or the number of iterations reaches the maximum number (set to 40).
\textbf{For KGE-TSP}, we set the embedding dimension to $500$. For HAKE-TSP, we select ${s}_{hrt}$ from $\{ 20, 10, 5, 3, 1,0.5, 0.1\}$. For PairRE-TSP, we select ${s}_{hrt}$ from $\{ 5000, 3000, 2500, 2000, 1000,500, 100\}$.
Adam \cite{adam} with an initial learning rate of $0.001$ is used for optimization. We adapt the learning rate by setting it to 80\% of the current learning rate if the loss does not decrease for $5$ steps.
We evaluate the model on valid data per 10000 steps.
\textbf{For GPHT}, in graph partition, at each time, we randomly extract $20$ subgraphs and select the most balanced one to store as a subgraph. 
We select $L$ from $ \{2, 3\}$ 
and ${\theta}_{hrt}\in \{ 5, 3, 1, 0.5, 0.1, 0.05, 0.01\}$ for GPHT(HAKE), ${\theta}_{hrt}\in \{ 100, 50, 30, 20, 10, 5, 1\}$ for GPHT(PairRE). 
The number of CompGCN layers is 1. 
We randomly select 20\% triples as the query set $\mathcal{T}^{que}$ in each subgraph. The model is optimized by Adam with learning rate as $3\times 10^{-5}$. 
Per $10$ times of training on all subgraphs,  we evaluate the model on valid data.

\begin{table*}[]
    \centering
    \caption{Averaged TSP results on $Wiki79k$ and $Wiki143k$ dataset evaluated under the RS-POWA from 3 times of experiments.}
    \resizebox{0.85\textwidth}{!}{
    \begin{tabular}{c| c | c c  c| c c >{\columncolor{light-gray}}c |>{\columncolor{light-gray}}c}
         \toprule
         \multirow{2}{*}{Datasets} & \multirow{2}{*}{Models} & \multicolumn{3}{c|}{Number of Triples in} &\multicolumn{3}{c|}{Classification Metrics} &  \cellcolor{white}Ranking Metric\\
         \cmidrule{3-9}
         & & $\mathcal{T}_{predict}$ & $\mathcal{T}_{predict}^{POWA}$  & $\mathcal{T}_{predict}^{POWA+}$ & $JPrecision$ & $STRecall$  & $F_{TSP}$ &  $RS_{TSP}$\\
         \midrule
         \multirow{5}{*}{Wiki79k} & RuleTensor-TSP &9188$\pm$0 & 1319$\pm$0 & 1292$\pm$0 & 0.560$\pm$0 & 0.128$\pm$0 & 0.208$\pm$0 & 0.347$\pm$0 \\
         & HAKE-TSP & 125$\pm$87 & 119$\pm$83 & 30$\pm$7 & 0.246$\pm$14.8\% & 0.044$\pm$0.4\% &0.075$\pm$0.3\% & 3.28$\pm$42\%\\
         & PairRE-TSP & 29742$\pm$105& 10217$\pm$51 & 2901$\pm$7 & 0.191$\pm$0.1\% & \textbf{0.428$\pm$0.0\%} &0.264$\pm$0.2\% & 3.18$\pm$7.4\%\\
         % & \new{RGCN-TSP} \\ 
         % & \new{CompGCN-TSP} \\
         \cmidrule{2-9}
         & GPHT(HAKE) & 209$\pm$47 & 191$\pm$47 & 44$\pm$8 & 0.220$\pm$8\% & 0.053$\pm$0.4\% &0.085$\pm$1.1\% & 3.29$\pm$46.6\%\\
         & GPHT(PairRE) & 12392$\pm$3813 & 5866$\pm$1262 & 2018$\pm$332 & \textbf{0.253$\pm$1.6\%} & 0.357$\pm$2.8\% & \textbf{0.296$\pm$0.4\%} & \textbf{3.92$\pm$40.5\%}\\
         \midrule
          \multirow{5}{*}{Wiki143k} & RuleTensor-TSP &24392 & 2570 & 2299& 0.494& 0.127 & 0.201 & 0.350  \\
         & HAKE-TSP & 22215$\pm$1283 & 6182$\pm$43 & 3044$\pm$72 & 0.315$\pm$0.2\% &\textbf{0.326$\pm$0.3\%} & 0.32$\pm$0.3\% & 5.33$\pm$15.2\% \\
         & PairRE-TSP & 19228$\pm$2075 & 3313$\pm$722 & 1191$\pm$77 & 0.211$\pm$3\% & 0.204$\pm$0.7\% & 0.207$\pm$1\% & 3.68$\pm$10.3\%\\
         % & \new{RGCN-TSP} \\ 
         % & \new{CompGCN-TSP} \\
         \cmidrule{2-9}
         & GPHT(HAKE) & 17702$\pm$7935 & 4709$\pm$60 & 2700$\pm$681 & 0.363$\pm$4.8\% & 0.307$\pm$4.2\% & \textbf{0.333$\pm$4.6\%} & \textbf{5.36$\pm$6.5\%} \\
         & GPHT(PairRE) & 3011$\pm$233 & 1954$\pm$191 & 909$\pm$34 & \textbf{0.384$\pm$1.9\%} & 0.178$\pm$0.3\% & 0.243$\pm$0.1\% & 5.04$\pm$16\% \\
         \bottomrule
    \end{tabular}
    }
    \label{tab:wiki}
\end{table*}

\begin{table*}[]
    \centering
    \caption{Averaged TSP results on CFamily dataset evaluated under the CWA from 3 times of experiments.}
    \resizebox{0.8\textwidth}{!}{
    \begin{tabular}{c | c c c | c c >{\columncolor{light-gray}}c| >{\columncolor{light-gray}}c }
         \toprule
     \multirow{2}{*}{Models} & \multicolumn{3}{c|}{Number of Triples in} &\multicolumn{3}{c|}{Classification Metrics} &  \cellcolor{white}Ranking Metric \\
         \cmidrule{2-8}
         & $\mathcal{T}_{predict}$ & $\mathcal{T}_{predict}^{CWA}$  & $\mathcal{T}_{predict}^{CWA+}$ & $JPrecision$ & $STRecall$  & $F_{TSP}$ &  $RS_{TSP}$\\
         \midrule
         RuleTensor-TSP & 911$\pm$0 & 911$\pm$0 & 572$\pm$0 & 0.628$\pm$0 & 0.158$\pm$0 & 0.252$\pm$0 & 1.99$\pm$0\\
         HAKE-TSP & 3186$\pm$543 & 3186$\pm$543& 1788$\pm$321& 0.561$\pm$1.8\% & 0.624$\pm$5.9\% & \textbf{0.591$\pm$3.1\%} & 6.37$\pm$11\%\\
         PairRE-TSP & 5732$\pm$2062 & 5732$\pm$2062 & 1747$\pm$694 & 0.305$\pm$1.8\% & 0.616$\pm$13.7\% & 0.408$\pm$4.9\% & 1.83$\pm$51\%\\
          \new{RGCN-TSP} & 30608$\pm$3443 & 30608$\pm$3443 & 3026$\pm$137 & 0.093$\pm$6.0\% & \textbf{0.704$\pm$0.5\%} & 0.163$\pm$5.2\% & -9.42$\pm$0.6\%\\ 
         \new{CompGCN-TSP} & 38931$\pm$3745 & 38931$\pm$3745 & 2599$\pm$47 & 0.062$\pm$4.8\% & 0.598$\pm$0.3\% & 0.114$\pm$1.7\% & -9.39$\pm$0.6\%\\
         \midrule
         GPHT(HAKE) & 1896$\pm$149 & 1896$\pm$149 & 1222$\pm$58 & \textbf{0.645$\pm$2.6\%} & 0.516$\pm$1.2\% & 0.573$\pm$0.4\%  &  \textbf{6.65$\pm$18\%}\\
         GPHT(PairRE) & 3739$\pm$593 & 3739$\pm$593 & 1471$\pm$187 & 0.393$\pm$1.3\% & 0.566$\pm$3.4\% &0.464$\pm$0.7\% & 3.16$\pm$31\%\\
         \bottomrule
    \end{tabular}
    }
    \label{tab:CFamily}
\end{table*}

\subsection{Incomplete KG Evaluation Under the RS-POWA}
\label{sec:evaluation-POWA}
We first conduct experiments on  Wiki79k and Wiki143k under the RS-POWA. 
Results are shown in Table \ref{tab:wiki}, in which we not only present the 4 metrics, but also show the number of triples in $\mathcal{T}_{predict}$, $\mathcal{T}_{predict}^{POWA}$ in Equation (\ref{equ:t_predicted_powa}), and $\mathcal{T}_{predict}^{POWA+}$ in Equation (\ref{equ:t_predicted_powa+}) that are used to calculate the metrics. 
% Among the metrics, 
We set the background of $F_{TSP}$ and $RS_{TSP}$ as grey since they are more important. $F_{TSP}$ is a combined metric of $JPrecision$ and $STRecall$. $RS_{TSP}$ is the only metric considering the ranking order of the triples in the predicted set. 

Firstly, let's have a look at the results of evaluation metrics. On Wiki79k, GPHT(PairRE) performs the best. It achieves the highest score on $JPrecision$, $F_{TSP}$ and $RS_{TSP}$ ($0.253$, $0.296$, and $3.92$, respectively). While on $STRecall$, PairRE-TSP performs better. 
On Wiki143k, GPHT(HAKE) performs the best on the most important two metrics $F_{TSP}$ and $RS_{TSP}$. GPHT(PairRE) and HAKE-TSP perform the best on $JPrecision$ and $STRecall$. Based on these results, we could conclude that (1) overall $GPHT$ performs better than baseline RuleTensor-TSP, HAKE-TSP, and PairRE-TSP; (2) the applied KGE in GPHT significantly affects the results. Specifically, if a KGE with KGE-TSP achieves higher TSP results, its corresponding GPHT(KGE) also performs better. 

Secondly, on Wiki79k, the largest and smallest predicted sets are from PairRE-TSP and HAKE-TSP, which include $29742$ and $125$ triples. Though a large predicted set is prone to have a higher $STRecall$, none of them performs the best on $JPrecision$, $T_{TSP}$ and $RS_{TSP}$. 
On Wiki143k, the largest and smallest predicted sets are from RuleTensor-TSP and GPHT(PairRE). Though a small predicted set is prone to have a higher $JPrecision$, none of them performs the best on $TRecall$, $F_{TSP}$ and $RS_{TSP}$.
This shows the reasonability of the evaluation metrics especially $F_{TSP}$ and $RS_{TSP}$
since a good result cannot be trickily achieved by predicting an extremely large or small triple set. 

Thirdly, let's have a look at the fluctuation of the results. The main results in Table \ref{tab:wiki} are averaged from experiments running 3 times. We also show the exact fluctuation of triple numbers in the $\mathcal{T}_{predict}, \mathcal{T}_{predict}^{POWA}$ and $\mathcal{T}_{predict}^{POWA+}$, and the fluctuation percentage compared to the main results for the classification and ranking metrics. Among all the methods, the most stable one is RuleTensor-TSP which always performs the same with the same hyperparameters. While for HAKE-TSP, PairRE-TSP, GPHT(HAKE), and PairRE(HAKE), the learning of which involves random factors, their performance varies when running the experiments multiple times. And sometimes, the fluctuation is significant, especially for the ranking metric. For example, in Wiki79k, there are $\pm42\%$, $\pm46.4\%$ and $\pm40.5\%$ on the $RS_{TSP}$ with HAKE-TSP, GPHT(HAKE), and GPHT(PairRE), while the fluctuation on $F_{TSP}$ is only $\pm0.3\%$, $\pm1.1\%$, and $\pm0.4\%$ correspondingly. 
This indicates that though the predicted triple sets from those methods of different runs are quite stable, the ranking order of the triples in the set may vary significantly, showing that the $RS_{TSP}$ is a more challenging metric than the classification metrics.

\begin{figure*}
    \centering
    \includegraphics[width=0.9\textwidth]{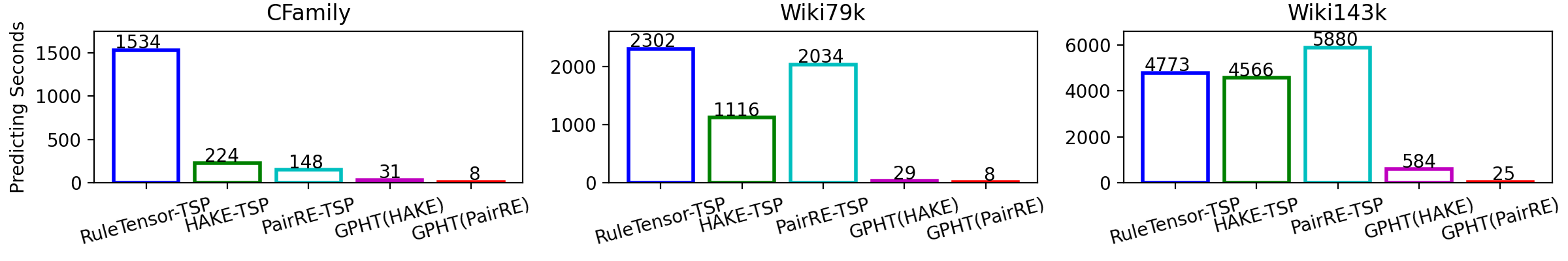}
    \vspace{-3mm}
    \caption{Predicting time of all methods on CFamily, Wiki79k, and Wiki143k.}
    \vspace{-3mm}
    \label{fig:time-analysis}
\end{figure*}

\begin{figure*}
    \centering
    \includegraphics[width=0.92\textwidth]{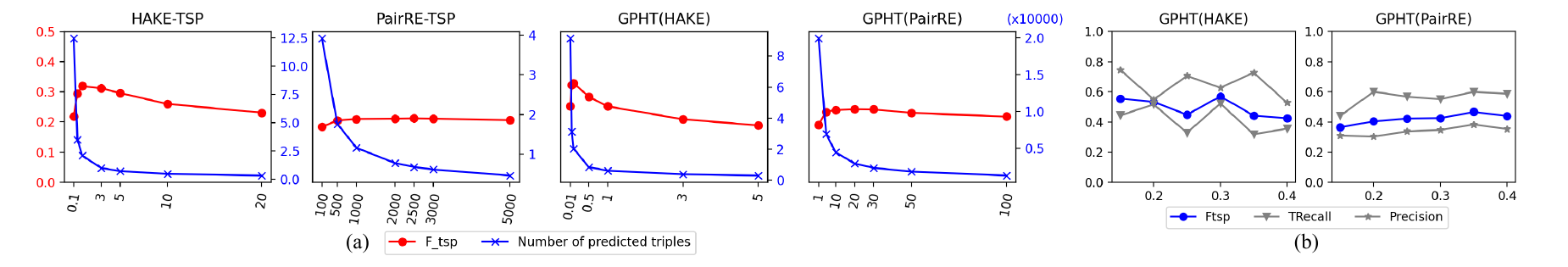}
    \vspace{-5mm}
    \caption{(a) Hyperparameter $\theta_{hrt}$ analysis, where the x axis in each figure is the $\theta_{hrt}$. \new{(b) Hyperparameter $\theta_{ht}$ analysis of the GPHT method.}}
    \vspace{-6mm}
    \label{fig:hrt-analysis}
\end{figure*}

\begin{figure}
    \centering
    \includegraphics[width=0.46\textwidth]{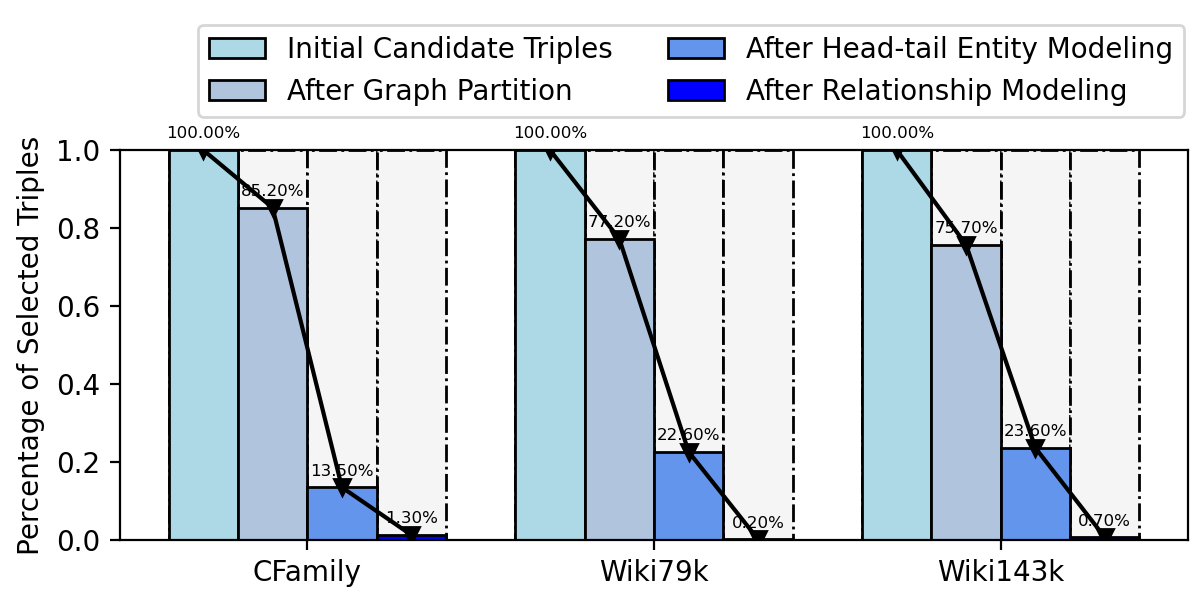}
    \vspace{-0.2cm}
    \caption{Percentage of triples left in different stage of the GPHT(HAKE).}
    \vspace{-3mm}
    \label{fig:triple-percentage}
\end{figure}

\subsection{Complete KG Evaluation Under the CWA}
Apart from RS-POWA applied to the realistic datasets, we conduct an experiment on the CFamily dataset under CWA, where $\mathcal{T}_{predict}=\mathcal{T}_{predict}^{CWA}$. Compared to the results evaluated under RS-POWA, the results of $JPrecision$, $F_{TSP}$ and $RS_{TSP}$ are different under CWA, and the result of $STRecall$ is the same. 
As shown in Table \ref{tab:CFamily}, the best performance on $STRecall$ and $F_{TSP}$ are achieved by HAKE-TSP, and the best performance on $JPrecision$ and $RS_{TSP}$ is achieved by GPHT(HAKE). 
This indicates HAKE-TSP gives better results from the perspective of classification and GPHE(HAKE) gives better results from the perspective of ranking, and overall methods with HAKE performs better than PairRE on CFamily. 
\new{
We also show RGCN-TSP and CompGCN-TSP results by adapting the RGCN \cite{rgcn} and CompGCN \cite{compgcn} to TSP task. Though they perform good on link prediction task, they perform poor on TSP task. This is because in RGCN and CompGCN, there are negative triple scores significantly higher than most positive triples. 
}
Similar to the results in Table \ref{tab:wiki}, neither the method predicting the largest triple set nor the method predicting the smallest triple set achieves the best results. And the results fluctuation is more significant on the ranking metrics.

\subsection{Efficiency Analysis}
One of the key challenges for TSP is the huge candidate triple space, which is closely related to the predicting time. 
In Figure \ref{fig:time-analysis}, we show the predicting time, from which we can see that different methods vary significantly. 
On CFamily, Wiki79k, and Wiki143k, the longest predicting time are 1534, 2302, and 5580 seconds, from RuleTensor-TSP, RuleTensor-TSP, and PairRE-TSP respectively. The overall predicting time of RuleTensor-TSP is relatively long, since it has to iteratively predict new triples via tensor calculation until it meets the stop conditions. The predicting time of HAKE-TSP and PairRE-TSP increases when the dataset size increases. Among all the methods, the predicting time of GPHT(HAKE) and GPHT(PairRE) is significantly shorter. This is because GPHT has effective candidate triple reduction strategies. To prove this, we show the percentage of triples left in different steps of GPHT in Figure \ref{fig:triple-percentage}. 
Before predicting, the initial candidate triples is regarded as $100\%$. 
After graph partition, $85.2\%$, $77.2\%$, and $75.7\%$ of the original triples are kept in CFamily, Wiki79k and Wiki143k. 
After head-tail entity modeling, i.e. keeping candidate triples with head and tail entity predicted to have missing relations by HTEM module, the number of candidate triples is further reduced to $13.5\%$, $22.5\%$ and $23.5\%$ of the original triples. 
And finally, with relation modeling, $1.3\%$, $0.2\%$ and $0.7\%$ of the original triples are output as the predicted triple set. 
Considering that the GPHT method achieves good performance on the TSP evaluation metrics, Figure \ref{fig:triple-percentage} strongly demonstrates the effectiveness of graph partition, the HTEM, and relationship modeling in reducing the candidate triple space. 
These steps could filter the false and keep the true candidate triples. And they cost less than calculating the score for each filtered triple. 
Thus GPHT's prediction time is significantly reduced.
\new{
Table \ref{tab:graph-partition-analysis} presents the statistics of subgraphs. As we can see, the denser the subgraph is, the higher percentage of head-tail pairs are predicted in HTEM. If two subgraphs with a similar density, the bigger one has a higher head-tail predicted percentage.}

\begin{table}[]
    \centering
    \caption{\new{Statistics of $\mathcal{G}_{part}$, including the number of subgraphs (\# $\mathcal{G}_i$), entities (\#E), relations (\#R), triple (\#T), density (Den), and predicted $(h,t)$ pair percentage (ht\%). }}
    \vspace{-3mm}
    \resizebox{0.49\textwidth}{!}{
    \begin{tabular}{l|c | c c c c c | c c c c c}
    \toprule
    
        & \multirow{2}{*}{\# $\mathcal{G}_i$} &  \multicolumn{5}{c|}{Largest $\mathcal{G}_i$} & \multicolumn{5}{c}{Smallest $\mathcal{G}_i$} \\ 
        & & \#E & \#R & \#T & Den & ht\% & \#E & \#R & \#T & Den &ht\% \\
        \midrule
        Wiki79k & 423 & 896 & 51  & 6829 & 0.85\% & 6.72\%
        % 53930(6.72\%) 
        & 36 & 18 & 18 &1.39\% 
        % & 48(3.70\%) 
        & 3.70\%
        \\
        Wiki143k & 371 & 2619 & 80 & 15013 & 0.22\% 
        % & 239120(3.49\%) 
        & 3.49\% 
        & 346 & 173 & 173 & 0.14\% 
        % & 289 (0.24\%) 
        & 0.24\%
        \\
        CFamily & 75 & 139 & 12 & 1257 & 6.50\% 
        % & 19321(1\%) 
        & 1.00\%
        & 154 & 77 & 77 & 0.32\% 
        % & 729 (3.07\%) 
        & 3.07\%
        \\
    \bottomrule
    \end{tabular}
    }
    
    \label{tab:graph-partition-analysis}
\end{table}
\vspace{-3mm}

\begin{table*}[]
    \centering
        \caption{Ablation study.}
        \vspace{-3mm}
        \resizebox{0.9\textwidth}{!}{
    \begin{tabular}{c l | c c c c | c c c c }
    \toprule
    & & \multicolumn{4}{c|}{GPHT(HAKE)} & \multicolumn{4}{c}{GPHT(PairRE)} \\
     &  & with all & -$\mathbf{a}_{ht}$ &  -$\mathbf{s}_{ht}$ & -$\mathbf{a}_{ht}$ -$\mathbf{s}_{ht}$ & with all & -$\mathbf{a}_{ht}$ &  -$\mathbf{s}_{ht}$ & -$\mathbf{a}_{ht}$ -$\mathbf{s}_{ht}$\\
     \hline
    % \multirow{4}{*}{ }&$Precision$ &  0.645$\pm$2.6\% & 0.584$\pm$6.4\%  & 0.630$\pm$9.0\% &  \textbf{0.660$\pm$2.9\%} &  \textbf{0.393$\pm$1.3\%} & 0.331$\pm$1.2\%  & 0.256$\pm$4.2\% & 0.345$\pm$2.3\%
      % \\ 
      % & $TRcall$ &  0.516$\pm$1.2\% &  \textbf{0.545$\pm$6.1\%} & 0.520$\pm$3.5\% & 0.508$\pm$5.8\%  &  \textbf{0.566$\pm$3.4\%} & 0.447$\pm$0.4\% & 0.414$\pm$3.8\%  & 0.545$\pm$1.2\%
      % \\
      % \rowcolor{light-gray}
     \cellcolor{white} CFamily  & $F_{TSP}$  & 0.573$\pm$0.4\%  & 0.564$\pm$0.8\%  & 0.570$\pm$2.5\% &  \textbf{0.574$\pm$2.2\%} & \textbf{0.464$\pm$0.7\%} & 0.380$\pm$0.6\%  & 0.316$\pm$3.7\%  & 0.423$\pm$1.3\%
      \\
      % \rowcolor{light-gray}
     \cellcolor{white}& $RS_{TSP}$ &  \textbf{6.65$\pm$18\%} & 6.11$\pm$9.0\%  & 6.43$\pm$20.0\%  & 6.44$\pm$17.0\% & 3.16$\pm$31.0\%& 1.65$\pm$62.0\%  &  2.42$\pm$45.0\% &  \textbf{3.41$\pm$37.0\%}
      \\
      \midrule
      %  \multirow{4}{*}{ } &$Precision$ &0.220$\pm$8\% & 0.344$\pm$46.2\%  & 0.429$\pm$38.8\%  &  \textbf{0.375$\pm$41.3\%} & 0.253$\pm$1.6\% & 0.245$\pm$4.7\%  & 0.217$\pm$0.8\% &  \textbf{0.263$\pm$0.2\%}
      %  \\ 
      % & $TRcall$ &  \textbf{0.053$\pm$0.4\%}  & 0.047$\pm$1.2\%  & 0.052$\pm$2.9\% & 0.044$\pm$11.8\% & 0.357$\pm$2.8\% &  0.382$\pm$9.5\%  &  \textbf{0.406$\pm$0.7\%} & 0.332$\pm$0.8\%
      % \\
      % \rowcolor{light-gray}
      \cellcolor{white} Wiki79k & $F_{TSP}$ &0.085$\pm$1.1\%  & 0.082$\pm$4.9\%  &  \textbf{0.094$\pm$0.6\%}   & 0.079$\pm$16.6\% & 0.296$\pm$0.4\% &  \textbf{0.298$\pm$0.4\%} & 0.283$\pm$0.5\% & 0.293$\pm$0.4\%
      \\
      % \rowcolor{light-gray}
      \cellcolor{white} & $RS_{TSP}$ & 3.29$\pm$46.6\%   & 2.26$\pm$46.0\%  &  \textbf{3.56$\pm$9.0\%}  & 2.50$\pm$88.0\%  & 3.92$\pm$40.5\%  & 
        \textbf{4.2$\pm$4.0\%}  & 3.92$\pm$38.0\% &  2.7$\pm$69.0\%
       \\
      \midrule
      %  \multirow{4}{*}{ } &$Precision$ & 0.363$\pm$4.8\% & 0.320$\pm$10.5\% &  \textbf{0.410$\pm$4.4\%}  &  \textbf{0.410$\pm$0.9\%} &  \textbf{0.384$\pm$1.9\%} &  0.189$\pm$0.8\%  & 0.326$\pm$3.2\%  & 0.355$\pm$14.3\%
      %  \\ 
      % & $TRcall$  & 0.307$\pm$4.2\%  &  \textbf{0.339$\pm$0.5\% } & 0.264$\pm$14.3\%   & 0.258$\pm$1.0\%  & 0.178$\pm$0.3\% &  \textbf{0.182$\pm$0.6\%}  &  \textbf{0.182$\pm$1.6\%} & 0.154$\pm$6.4\%
      % \\
      % \rowcolor{light-gray}
      \cellcolor{white}Wiki143k & $F_{TSP}$ &  \textbf{0.333$\pm$4.6\%} & 0.330$\pm$4.9\%  & 0.321$\pm$10.3\%   & 0.317$\pm$0.3\%  &  \textbf{0.243$\pm$0.1\%}  & 0.185$\pm$0.1\%  & 0.234$\pm$0.3\%  & 0.215$\pm$1.5\% \\
      % \rowcolor{light-gray}
      \cellcolor{white} & $RS_{TSP}$  &  \textbf{5.36$\pm$6.5\%}  & 4.3$\pm$42.0\%  & 4.52$\pm$118\%  & 3.86$\pm$29.0\%  &  \textbf{5.04$\pm$16\%} & 3.57$\pm$13.0\% & 4.44$\pm$26.0\% & 4.24$\pm$7.0\% \\

         \bottomrule
         
    \end{tabular}
    }

    \vspace{-3mm}

    \label{tab:ablation}
\end{table*}

\subsection{Hyperparmeter Analysis}

$\theta_{hrt}$ is a hyperparameter for selecting the final predicted triples. 
In Fig. \new{\ref{fig:hrt-analysis} (a)}, we show how the number of predicted triples and the evaluation metric $F_{TSP}$ vary with $\theta_{hrt}$ set from small to large for GPHT(HAKE) and GPHT(PairRE). We also show the figures for KGE-TSP, since set up $\theta_{hrt}$ is also used in the KGE-TSP. 
As we can see, for these four methods, when the threshold becomes larger, fewer triples are included in the final predicted set, and the $F_{TSP}$ firstly increases and then decreases. \new{When applying GPHT, 
we suggest starting with large variations in $\theta_{hrt}$ to identify the range where results initially increase and then decrease.}
Comparing the KGE-TSP and GPHT(KGE) with the same KGE, we can see that the threshold range is much smaller in GPHT(KGE) than in HAKE-KGE. 
Thus it is easier to set a proper $\theta_{hrt}$ in GPHT(KGE) during experiment. 
In our opinion, it is because the range of the truth value of the candidate triples input to the relationship modeling module in GPHT(KGE) is much smaller than all the candidate triples in KGE-TSP.
And we also observe that with the same $F_{TSP}$ result, GPHT(KGE) usually outputs fewer triples than KGE-TSP. This means the GPHT could achieve a comparable TSP result with a smaller predicted triple set than KGE-TSP. This is a good and expected property of triple set prediction method. 

$\theta_{ht}$ is a hyperparameter for selecting the head-tail entity pairs that are likely to have missing relations in the HTEM module. In Fig. \new{\ref{fig:hrt-analysis} (b)}, we show how the $F_{TSP}$, $STRecall$ and $JPrecision$ vary in different $\theta_{ht}$ settings. For GPHT(HAKE), when we increase $\theta_{ht}$ from $0.1$ to $0.4$, there are performance oscillations, and the best result is achieved by $\theta_{ht} = 0.3$. For GPHT(PairRE), when we increase $\theta_{ht}$, the performance slowly increases and then decreases. The best result is achieved by $\theta_{ht}=0.35$. 
Based on these results, we recommend setting $\theta_{ht} = 0.3$ as the first trial of hyperparameter search of  $\theta_{ht}$ when applying GPHT.
The difference in performance trend for GPHT(HAKE) and GPHT(PairRE) demonstrates, again, that our GPHT method relies on the applied KGE method and could benefit from the development of KGE in the future.

\subsection{Ablation Study}
In the HTEM module, we design the entity attention score $\mathbf{a}_{ht}$ and relation attention vector $\mathbf{s}_{ht}$ to help head-tail entity pair selection. In Table \ref{tab:ablation}, we show the ablation study on removing $\mathbf{a}_{ht}$, removing $\mathbf{s}_{ht}$ and removing both, i.e. columns of $-\mathbf{a}_{ht}$, $-\mathbf{s}_{ht}$ and $-\mathbf{a}_{ht}-\mathbf{s}_{ht}$, respectively. On CFamily, with both of them, GPHT(HAKE) achieves the best $RS_{TSP}$ result and GPHT(PairRE) achieves the best $F_{TSP}$ result. On Wiki79K, GPHT(HAKE) with $-\mathbf{s}_{ht}$ performs the best and GPHT(PairRE) with $-\mathbf{s}_{ht}$ performs the best. On Wiki143k, the GPHT(HAKE) and GPHT(PairRE) perform best including both $\mathbf{a}_{ht}$ and $\mathbf{s}_{ht}$. Though with $\mathbf{a}_{ht}$ and $\mathbf{s}_{ht}$, the GPHT method does not always achieve the best performance compared to the ablation setting, but it performs better in 6 of 12 experiments. This demonstrates the effectiveness of  $\mathbf{a}_{ht}$ and $\mathbf{s}_{ht}$. 
From Table \ref{tab:ablation}, we can see that sometimes applying one of $\mathbf{a}_{ht}$ and $\mathbf{s}_{ht}$ gives better results, thus 
a more dynamic combination of $\mathbf{a}_{ht}$ and $\mathbf{s}_{ht}$ might introduce more robust performance, which will be investigated in our future work.

\section{Related Work}
Three are three common KG Construction methodologies:
manual construction \cite{gene2004gene}, automatic extraction from text \cite{al2020named}, and inference based on existing triples \cite{chen2020knowledge}. 
In this work, we define TSP as inferring new triples based on existing ones. Thus we mainly introduce existing KGC methods learned from existing triples, considering three major types. 

\paragraph{Rule-based methods} Rule learning and inference is a classical way for KGC. Rule learning aims to learn the inference rules $head \gets body$ that could be used to infer new triples, in which rule structure and quality learning are the key points. 
There are diverse forms of rules regarding their elements, such as path rules \cite{Ptranse}, horn rules \cite{galarraga2013amie}, rules with constants \cite{meilicke2019anytime}, and rules with negation \cite{ortona2018robust}. Some rule learning methods \cite{galarraga2013amie,meilicke2019anytime,ho2018rule, amie3} first search the structure of rules and then evaluate rule quality. During structure search, candidate rule extending and pruning strategies are applied to increase the diversity of rules and reduce the search space. During quality evaluation, quality metrics such as \textit{support}, \textit{confidence}, \textit{PCA confidence}, and \textit{head coverage} are used.  
These search-based methods are inefficient on large-scale KGs. Thus some methods combine rule learning with tensor calculation. For example, differentiable rule learning methods \cite{DBLP:conf/nips/YangYC17, DBLP:conf/nips/SadeghianADW19, DBLP:conf/iclr/WangSDK20, DBLP:journals/corr/abs-2209-05815} learn rule structure and rule quality at the same time through adapted Tensorlog \cite{tensorlog}, where entities and relations are represented as tensors. The inference process is modeled as tensor calculations. Some works \cite{DBLP:journals/tkde/OmranWW21,ho2018rule,omran2018scalable} integrate KG embeddings to search rule structures and use pruning to reduce the search space or overcome the negative impact of KG incompleteness. Some works \cite{rnnlogic, kang2020learning} regard rules as latent variables and train neural networks for learning and inference. \baserule~ proposed in this work is a combination of search-based rule mining and tensor-based rule inference.  

\paragraph{Embedding-based methods} KG embedding methods aim to embed KGs into a vector space.
The most widely chosen vector space is Euclidean space \cite{transe,distmult}, and other spaces such as complex \cite{complex,rotate}, quaternion \cite{DBLP:conf/nips/0007TYL19,DBLP:conf/aaai/CaoX0CH21}, geometric \cite{boxe,elembedding,cone}, hyperbolic \cite{DBLP:conf/acl/ChamiWJSRR20,DBLP:conf/emnlp/SunCHWDZ20}, probabilistic distribution \cite{dirie} space are also used. 
With selected vector space, entities are represented in the space, for example as a point in Euclidean space \cite{transe}, and relations are regarded as transformation functions defining how the head entity embedding could be transformed to the tail entity embedding under a certain relation. 
The transformation function can be 
addition \cite{transe}, multiplication \cite{distmult}, rotation \cite{distmult}, or neural networks \cite{conve} with parameters. 
During the design of vector space assumption, methods are expected to be able to model diverse types of knowledge, such as  N-N \cite{transh}, symmetry/antisymmetry \cite{paire,complex}, transitive \cite{dorc}, inverse \cite{rotate}, compositional \cite{rotate} relations,  entity hierarchy \cite{hake} and relation hierarchy \cite{boxe}.     
The training of KGE methods relies on sampling negative triples, whose quality significantly affects the training results. Typical negative sampling methods include sampling following uniform distribution \cite{transe}, Bernoulli distribution \cite{transh}, self-adversarial weights \cite{rotate}, adversarial learning \cite{DBLP:conf/naacl/CaiW18}, adaptive mixup \cite{DBLP:conf/semweb/ChenZYCT23} and so on. In this work, we apply HAKE \cite{hake} with self-adversarial negative sampling due to its efficiency and effectiveness. 
Most other KGE methods can also be applied. 

\paragraph{GNN-based methods} 
Relational GNNs are proposed to explicitly encode the graph structure of a KG, mostly following the encoder-decoder framework \cite{rgcn}. 
The first method is R-GCN \cite{rgcn}, which defines relation-specific aggregation functions to make the GNN relation-aware. 
CompGCN \cite{compgcn} defines a  variety of entity-relation  composition  operations to overcame the over-parameterization problem and updates the entity and relation embeddings in each CompGCN layer.
M-GNN \cite{mgnn} designs a powerful GNN layer using multi-layer perceptrons applied on a series of coarsened graphs created following graph coarsening schemes, to model hierarchical structures in KG. 
RGHAT \cite{rghat} is a relational GNN with hierarchical attentions which highlight the  importance  of  different neighbors of an entity.
SE-GNN \cite{segnn} is a semantic evidence aware GNN in which entity, relation, and triple level semantic evidence are considered, modeled, and merged by a multi-layer aggregation. We adopt CompGCN in this work due to its effectiveness, but other GNNs can also be applied.

\section{Conclusion} 
In this paper, we propose a new task named Triple Set Prediction (TSP) that allows automatic and end-to-end KG completion starting from zero.
For the new task, we propose three classification and a ranking evaluation metrics both considering  the open-world assumption applied in KG representation.
We propose a novel TSP method GPHT to handle the huge candidate triple space.
In order to compare the performance, we also adapt the rule-based and embedding-based KGC methods to TSP. We conduct extensive experiments on three datasets. Results demonstrate the possibility of predicting missing triple set from zero and the effectiveness of our GPHT method.  
In the future, we would like to explore better candidates triple space reducing strategy to make TSP more efficient and test TSP task on real-life applications.

\section*{Acknowledgments}
\new{We sincerely thank the valuable suggestions from the reviewers and editors on our work.} This work is funded by NSFC62306276, Zhejiang Provincial Natural Science Foundation of China (No. LQ23F020017) and Yongjiang Talent Introduction Programme (2022A-238-G), Ningbo NSF (2023J291) and NSFC91846204/U19B2027.

\bibliographystyle{IEEEtran}
\bibliography{main-reference}

\vspace{-20mm}

\begin{IEEEbiography}[{\includegraphics[width=1in,height=1.25in,keepaspectratio]{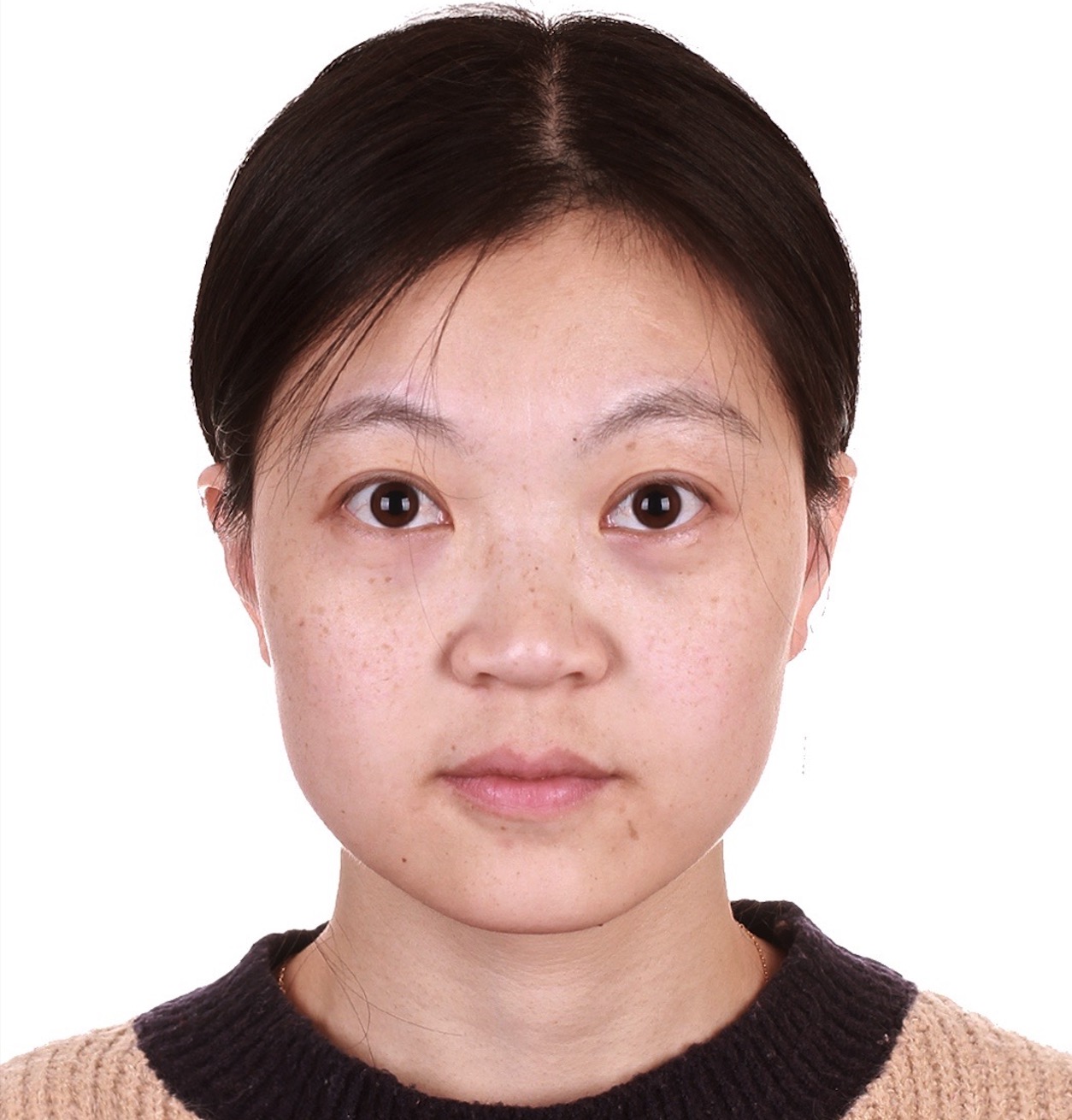}}]{Wen Zhang}
is an Asistant Professor at School of Software Technology in Zhejiang University.  
Her research interests are knowledge graph, knowledge representation and reasoning, and graph learning. 
\end{IEEEbiography}
\vspace{-15mm}

\begin{IEEEbiography}
[{\includegraphics[width=1in,height=1.25in,clip,keepaspectratio]{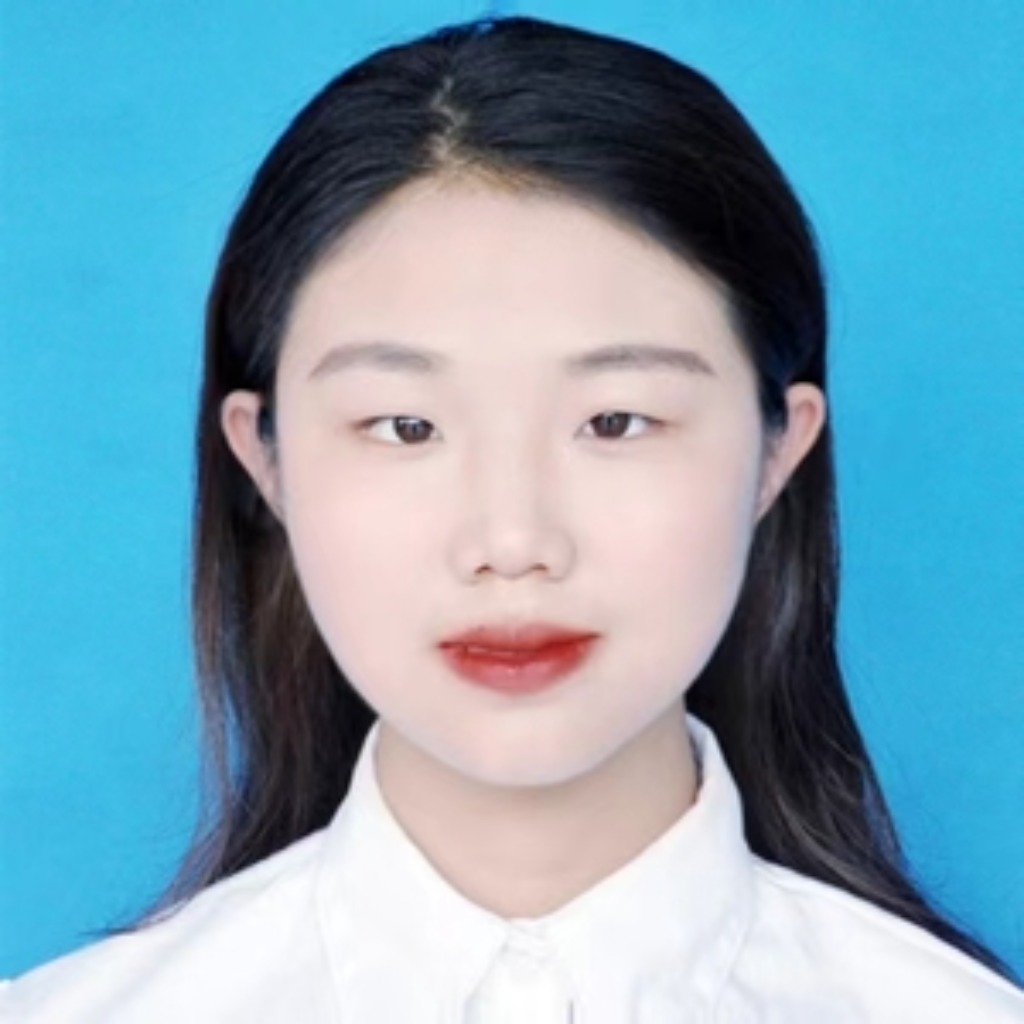}}]
{Yajing Xu}is currently pursuing the PhD degree with the School of Computer Science and Technology, Zhejiang University, China. Her research interests are knowledge graph  completion and multi-modal knowledge graph construction.
\end{IEEEbiography}
\vspace{-15mm}

\begin{IEEEbiography}
[{\includegraphics[width=1in,height=1.25in,clip,keepaspectratio]
{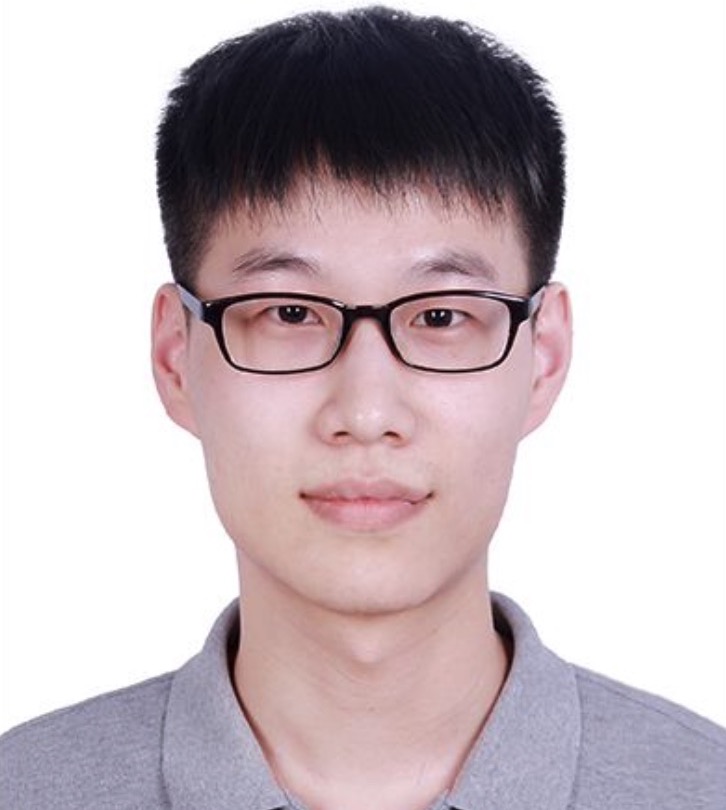}}]{Peng Ye} received the BEc degree from the Southeast University in 2019, the MEc degree from the Zhejiang University in 2023. He is currently work in China Mobile (Zhejiang) Innovation Research Institute Co., Ltd. His research interests include data mining and information retrieval, mainly focusing on knowledge graph completion and prediction. 
\end{IEEEbiography}
\vspace{-15mm}

\begin{IEEEbiography}
[{\includegraphics[width=1in,height=1.25in,clip,keepaspectratio]{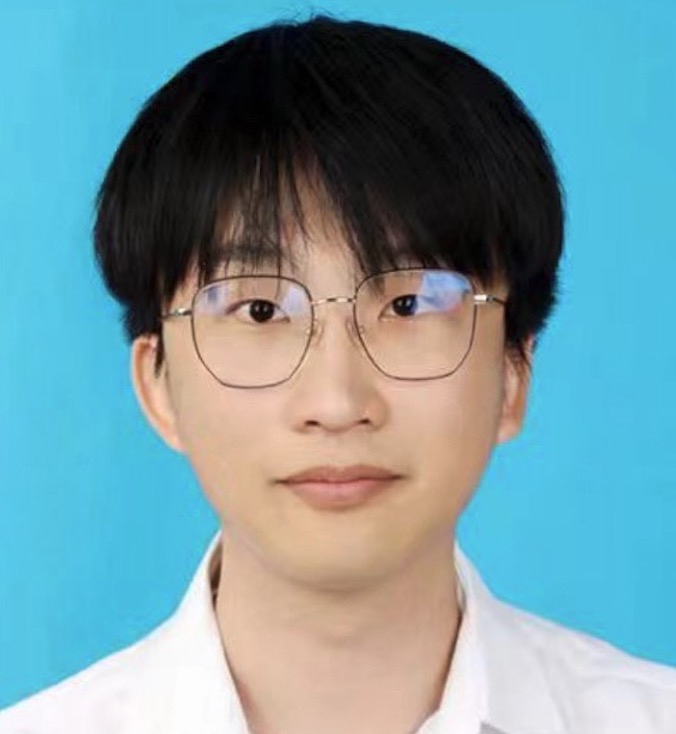}}]
{Zhiwei Huang} is currently pursuing a Master's degree in the School of Software at Zhejiang University, China. His research interests mainly focus on knowledge graph representation and reasoning.. 
\end{IEEEbiography}
\vspace{-15mm}

\begin{IEEEbiography}
[{\includegraphics[width=1in,height=1.25in,clip,keepaspectratio]
{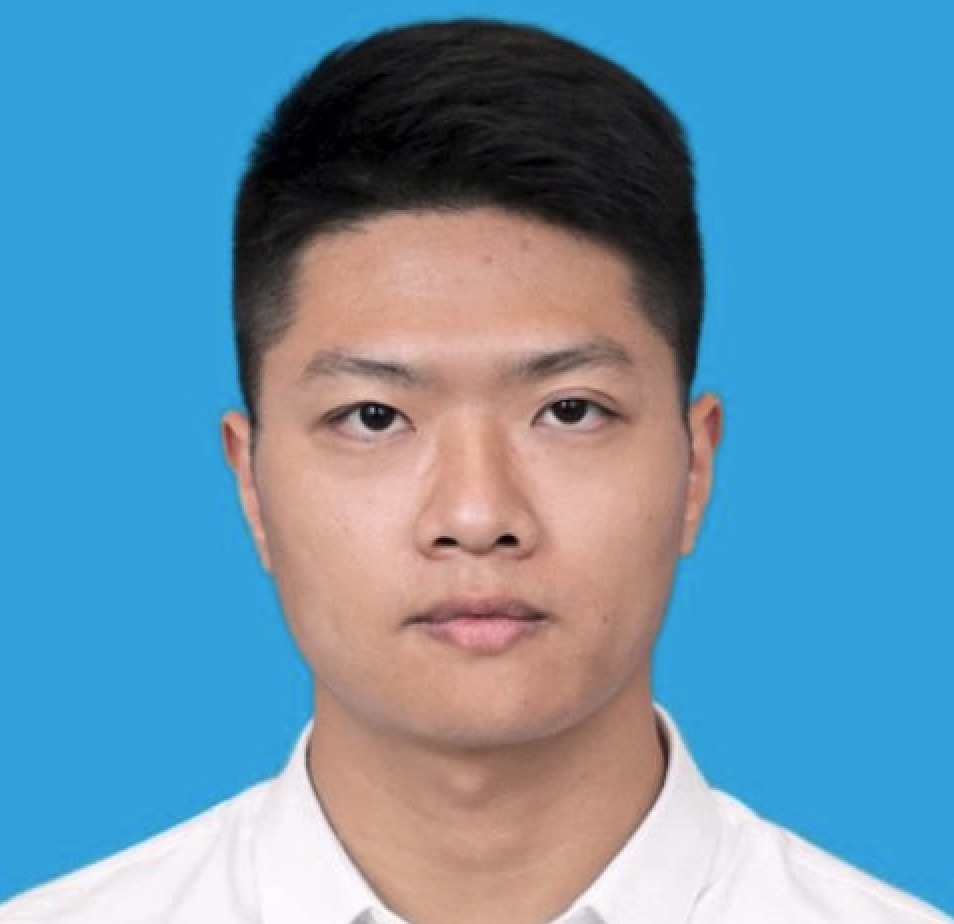}}]{Zezhong Xu} is currently pursuing the PhD degree with the School of Computer Science and Technology, Zhejiang University, China. His research interests mainly focuses on neural and symbolic reasoning, including rule mining and complex query answering on Knowledge graph. 
\end{IEEEbiography}
\vspace{-15mm}

\begin{IEEEbiography}
[{\includegraphics[width=1in,height=1.25in,clip,keepaspectratio]
{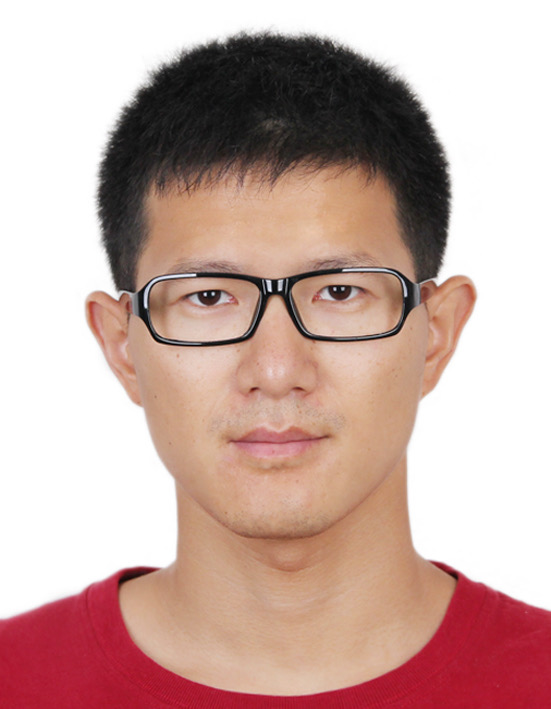}}]{Dr. Jiaoyan Chen } is a Lecturer (Assistant Professor) in Department of Computer Science, University of Manchester, and a part-time Senior Researcher in Department of Computer Science, University of Oxford. Dr. Chen does research and teaching mainly on Knowledge Graph, Ontology, Semantic Web and Machine Learning.
\end{IEEEbiography}
\vspace{-10mm}

\begin{IEEEbiography}[{\includegraphics[width=1in,height=1.25in,clip,keepaspectratio]{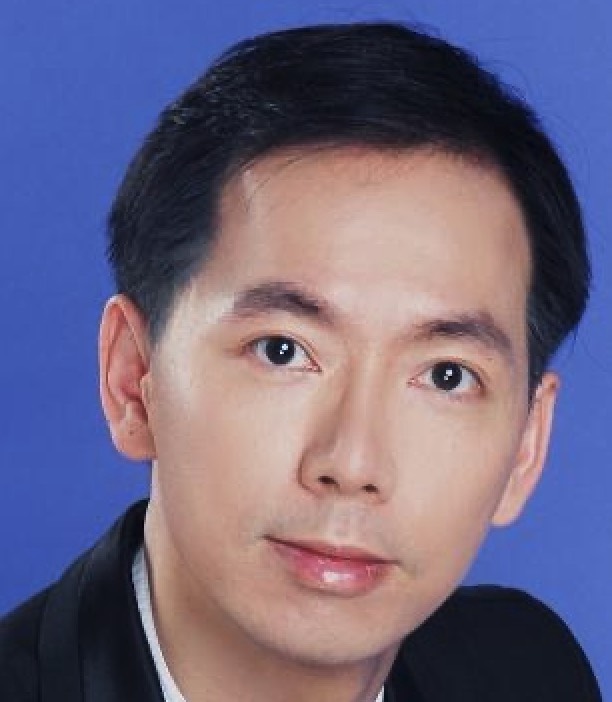}}]{Jeff Z. Pan } is the Reader in Knowledge Graphs of the School of Informatics in the University of Edinburgh. His research interests includes Knowledge representation and artificial intelligence, knowledge based reasoning and learning, knowledge based natural language understanding and generation.
\end{IEEEbiography}

\begin{IEEEbiography}[{\includegraphics[width=1in,height=1.25in,clip,keepaspectratio]{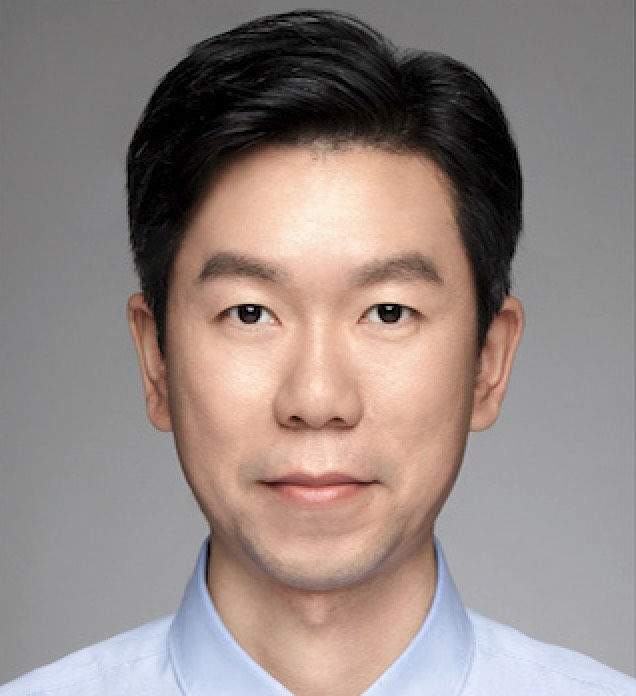}}]{Huajun Chen} a full professor of College of Compouter Science and Technologies at Zhejiang University, and a deputy director of the Key Lab of Big Data Intelligence at Zhejiang Province.  He received bachelor's degree and a PhD from Zhejiang University in 2000 and 2004 respectively. His research interests are Knowledge Graph and  Natural Language Processing, Big Data and  Artificial Intelligence.  
\end{IEEEbiography}

\vfill

\end{document}